# Supervised Machine Learning and Physics based Machine Learning approach for prediction of peak temperature distribution in Additive Friction Stir Deposition of Aluminium Alloy

Akshansh Mishra[1]

[1]School of Industrial and Information Engineering, Politecnico Di Milano, Milan, Italy

Mail id: akshansh.mishra@mail.polimi.it

**Abstract:** Additive friction stir deposition (AFSD) is a novel solid-state additive manufacturing technique that circumvents issues of porosity, cracking, and properties anisotropy that plague traditional powder bed fusion and directed energy deposition approaches. However, correlations between process parameters, thermal profiles, and resulting microstructure in AFSD remain poorly understood. This hinders process optimization for properties. This work employs a framework combining supervised machine learning (SML) and physics-informed neural networks (PINNs) to predict peak temperature distribution in AFSD from process parameters. Eight regression algorithms were implemented for SML modeling, while four PINNs leveraged governing equations for transport, wave propagation, heat transfer, and quantum mechanics. Across multiple statistical measures, ensemble techniques like gradient boosting proved superior for SML, with lowest MSE of 165.78. The integrated ML approach was also applied to classify deposition quality from process factors, with logistic regression delivering robust accuracy. By fusing data-driven learning and fundamental physics, this dual methodology provides comprehensive insights into tailoring microstructure through thermal management in AFSD. The work demonstrates the power of bridging statistical and physics-based modeling for elucidating AM process-property relationships.

## 1. Introduction

Additive manufacturing (AM) has revolutionized production across diverse industries by enabling on-demand fabrication of complex geometries directly from digital models [1-5]. While powder-bed fusion and directed energy deposition techniques dominate the metal AM landscape, they inherently suffer from issues related to the melt-pool mode of material addition and subsequent rapid solidification. Porosity, cracking, residual stresses, and anisotropic properties are common [6-9]. In recent years, solid-state metal AM approaches have emerged seeking to circumvent these challenges by avoiding bulk melting of feed material. Additive Friction Stir Deposition (AFSD) is one such novel technique combining concepts of friction stir processing and additive layer manufacturing. First proposed in 2018, it has garnered significant interest for promising superior microstructure and properties compared to other metal AM methods [10-15].

One of the primary benefits of AFSD is its ability to produce fully dense parts with properties that are comparable to those of wrought alloys. This is particularly important in industries such as aerospace, automotive, and defense, where high structural integrity and mechanical performance are crucial [16-19]. In addition, AFSD allows for localized deposition, enabling the precise replacement of lost material while maintaining the original properties of the component. This makes it an ideal solution for repairing and refurbishing high-value components that have been damaged due to wear, corrosion, or impact. Another significant advantage of AFSD is its capability for solid-state recycling of machining chips and scraps. By converting these materials into usable metal powder feedstock, AFSD enables sustainable in-house recycling, reducing waste and minimizing environmental impact. Moreover, AFSD offers higher deposition rates compared to other additive manufacturing techniques such as powder bed fusion and directed energy deposition, allowing for the rapid production of large parts. The unique deposition mechanism employed by AFSD also produces parts with improved microstructural properties. The as-deposited microstructure is characterized by fine-grained, equiaxed grains that result from dynamic recrystallization during the deposition process. These grains lead to isotropic properties that are superior to the coarse columnar grains found in other AM processes. Furthermore, the solid-state deposition process used in AFSD generates compressive residual stresses that enhance fatigue life and damage tolerance, whereas other AM methods typically induce tensile residual stresses.

The peak temperature distribution significantly influences microstructure evolution during AFSD. A comprehensive understanding of this relationship is necessary to predict and control microstructural features, such as grain size and shape, which directly impact the mechanical properties of the deposited material. High peak temperatures can generate residual stresses that may result in warpage or distortion of the workpiece [20-22]. By examining the correlation between peak temperature distribution and residual stress/distortion, this study aims to provide guidelines for mitigating these issues and ensuring dimensional accuracy. The heat affected zone (HAZ) resulting from AFSD can have a profound effect on the material's properties, including its microstructure, hardness, and corrosion resistance. Analyzing the peak temperature distribution within the HAZ can help researchers and manufacturers better understand its formation and develop strategies to minimize any detrimental effects. Uncontrolled peak temperatures can lead to thermal damage, altering the material's chemical composition, promoting phase transformations, or even causing burnout. By monitoring peak temperature distribution, this study seeks to identify threshold values that can prevent these unwanted phenomena and ensure the quality of the deposit. Different materials exhibit unique responses to the same processing conditions. Investigating peak temperature distribution across diverse materials can reveal material-specific trends and guide the development of tailored AFSD processes that account for these differences. Peak temperature distribution plays a crucial role in determining the optimal process parameters for achieving desired microstructures and mechanical properties. By analyzing the effects of processing conditions on peak temperature distribution, this study seeks to establish correlations that can facilitate the optimization of AFSD processes.

While additive manufacturing techniques like powder bed fusion and directed energy deposition are now commonplace, they suffer from issues like porosity, cracking, and poor mechanical properties due to the melt-pool mode of material addition. Recently, additive friction stir deposition (AFSD) has emerged as a novel solid-state technique combining friction stir processing and layer-wise deposition to circumvent these challenges. However, a comprehensive understanding of process-property-microstructure relationships in AFSD is lacking. In particular, the correlation between peak temperature distribution and resulting microstructural evolution during AFSD remains poorly understood. No prior work has systematically analyzed peak temperature profiles in AFSD to identify process-microstructure linkages. This knowledge gap hinders the optimization of AFSD processes for desired properties.

This research aims to address this gap by employing a dual supervised and physics-informed machine learning approach to predict peak temperature distribution from AFSD process parameters. By fusing data-driven learning and fundamental process physics, this work uniquely combines the strengths of both techniques to enhance model accuracy. The integrated framework is applied to modeling thermal phenomena in metal additive manufacturing. Outcomes from this data-driven and physics-based modeling will provide novel, clinically-relevant insights into tailoring microstructural features like grain morphology and size distribution through controlled thermal management. By clarifying peak temperature-property correlations, this work will facilitate the design of optimized AFSD processes for properties like strength and damage tolerance.

## 2. Comparison of Supervised ML and Physics based ML

Supervised machine learning and physics-based machine learning are two different approaches to building machine learning models. Training a model with labeled data in supervised machine learning entails knowing the target outcome for each input. In order for the model to make predictions on fresh, unforeseen data, it must be learned a mapping from inputs to outputs. The model's parameters are changed to reduce this loss after training using a loss function to measure the discrepancy between expected and actual results. Image classification, audio recognition, and sentiment analysis are a few examples of supervised machine learning tasks. On the other hand, physics-based machine learning entails including physical rules or restrictions in the machine learning model. This strategy is especially helpful when working with complicated systems whose basic physical principles are clear but whose behavior is unpredictable because of the sheer volume of data involved. To forecast the behavior of such systems and to make sure that the predictions are in line with the underlying physical rules, physics-based machine learning models can be used.

The discrepancy between the expected and actual output is frequently used to determine the loss function in supervised machine learning. The loss function in physics-based machine learning may contain terms that require the fulfilment of physical restrictions or rules, such as the conservation of energy or momentum as shown in Figure 1. While physics-based machine learning can work with unlabeled data and use the physical laws as a guide to discover the patterns in the data, supervised machine learning requires tagged training data. While physics-based machine learning models are created with the physical laws incorporated into the model architecture, supervised machine learning models are frequently created without any prior knowledge of the underlying physical principles. Unlike physics-based machine learning models, which may make probabilistic predictions that account for both measurement noise and the uncertainty in the physical laws, supervised machine learning models only make point predictions. Compared to supervised machine learning models, physics-based machine learning models are frequently easier to interpret because they shed light on the fundamental principles that underlie the system's behavior. In contrast to supervised machine learning models, physics-based machine learning models can be relatively cheaper to train and evaluate computationally, especially when dealing with complicated physical systems.

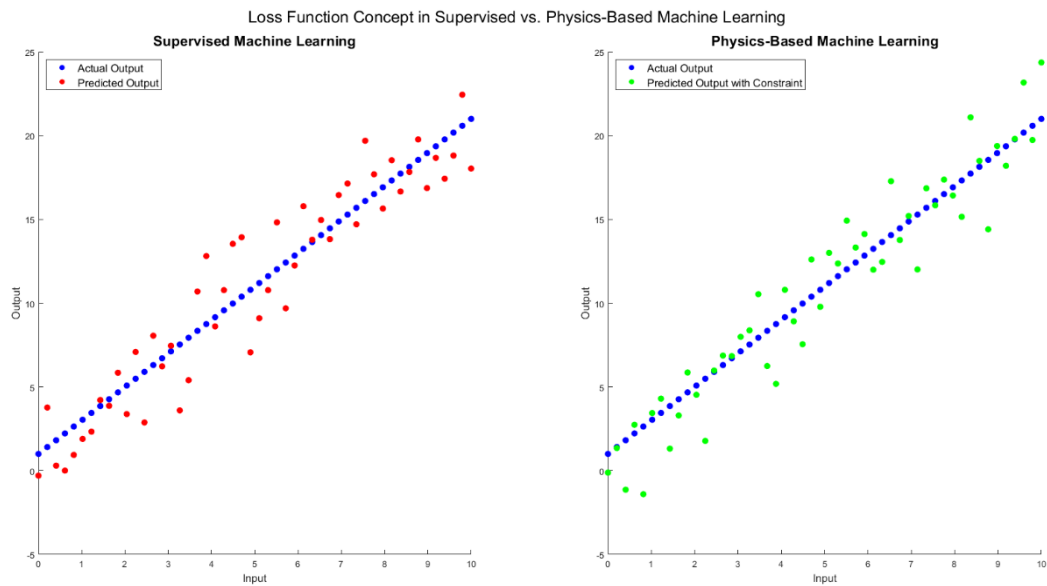

Figure 1. Concept of the Loss function in case of Supervised Machine Learning and Physics based Machine Learning approach

## 3. Materials and Methods

Additive Friction Stir Deposition (AFSD) is a solid-state additive manufacturing process that does not involve any melting of the material being deposited. It is based on the principles of friction stir welding and combines concepts of material deformation processing and layer-by-layer additive fabrication as shown in Figure 2. In AFSD, the material to be deposited is supplied in the form of a rod or powder into the rotating non-consumable tool. As the tool contacts the substrate or previous layer, friction at the interface generates heat which softens the feed material allowing it to plastically deform. For powder feedstocks, additional external heating may be required for proper consolidation. The rotating action of the tool provides mixing and consolidation of the feed material, which extrudes under pressure to fill the gap between the tool and substrate.

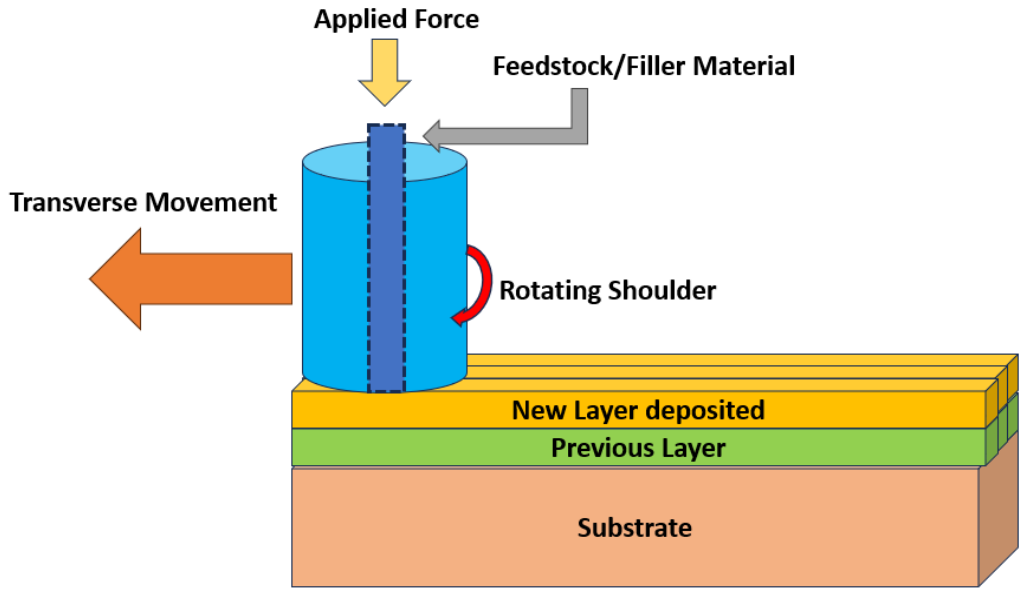

Figure 2. Schematic representation of Additive Friction Stir Deposition process

The severe plastic deformation occurring at elevated temperatures leads to metallurgical bonding between the deformed feed material and substrate material through interdiffusion at the interface. This bonding occurs entirely in the solid state unlike in melt-based processes. As the tool traverses across the substrate, it deposits a track of the solid-state deformed material. By repeating this process in a layer-by-layer fashion, complex 3D metal parts can be fabricated. The unique thermomechanical conditions of AFSD lead to dynamic recrystallization of the deposited material resulting in a fine equiaxed grain structure. This is unlike the coarse columnar grains commonly observed in melt-based additive manufacturing processes which suffer from problems like porosity, cracking, and anisotropic properties. The fine-grained microstructure achieved in AFSD enhances the material properties.

The main framework implemented in the present work is shown in Figure 3. The data [23-24] were prepared and were further imported to the Google Colab platform for subjecting it to machine learning algorithms coded using Python programming language. The initial input parameters considered in the present work are Rotational Rate (RPM)     Travel     Speed (mm/min), Tool Geometry, Deposition Material Flow Rate ($mm^3$/min), Tool Diameter (mm), and Powder Size (micro meter) while the output parameters are Peak temperature (degree Celsius) and the deposition quality. The data were further divided into training set and testing set i.e. 80 percent of the data were used for training purpose and 20 percent of the data were used for testing purpose. For predicting the peak temperature the data were subjected to eight supervised machine learning regression based algorithms and also to the four physics based machine learning algorithms. While for predicting the deposition quality, the data were subjected to nine classification based machine learning algorithms. In order to evaluate the performance for supervised regression and physics machine learning based models metric features such as MSE, MAE and R square value were used but for classification based models ROC-AUC Score and F1-score were used.

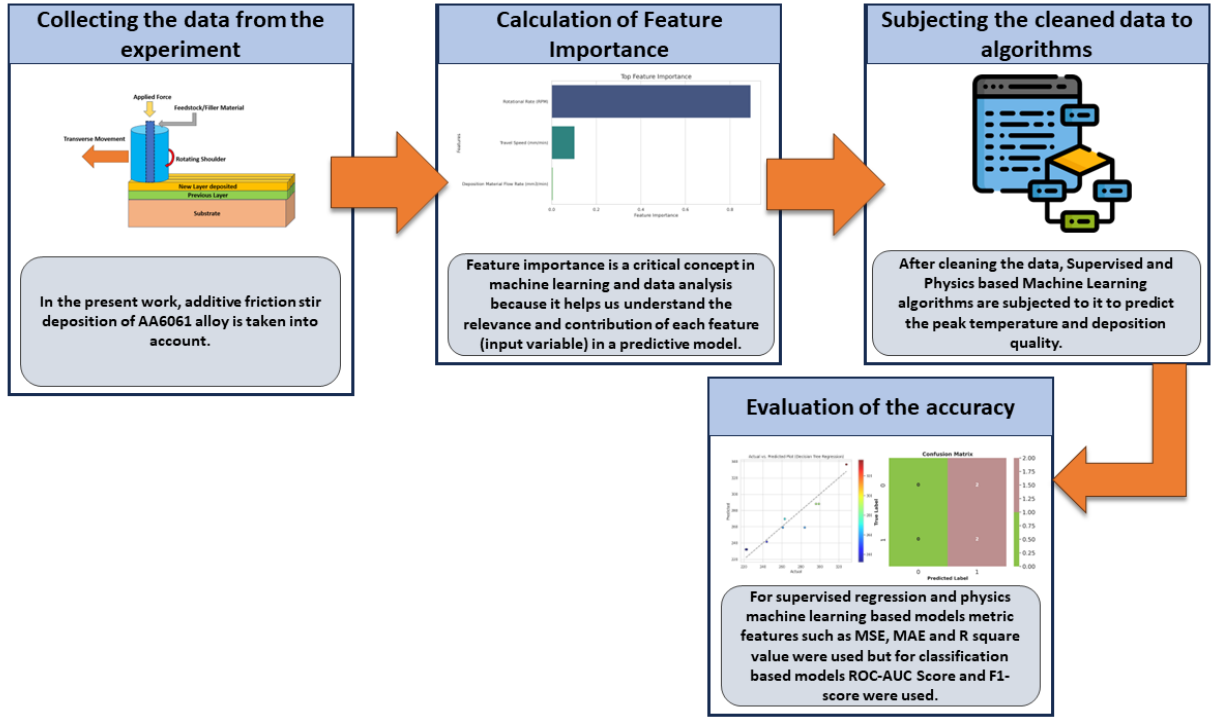

Figure 3. The machine learning framework implemented in the present work

## 4. Results and Discussion

### 4.1 Supervised Machine Learning Regression based algorithms for Peak temperature prediction

Figure 4 shows the correlation heatmap obtained for supervised machine learning regression based algorithms. We can see strong positive correlation between Rotational Rate (RPM) and Peak Temperature. There is also moderate positive correlation between Travel Speed and Peak Temperature. This indicates RPM and Travel Speed are important predictors of Peak Temperature.

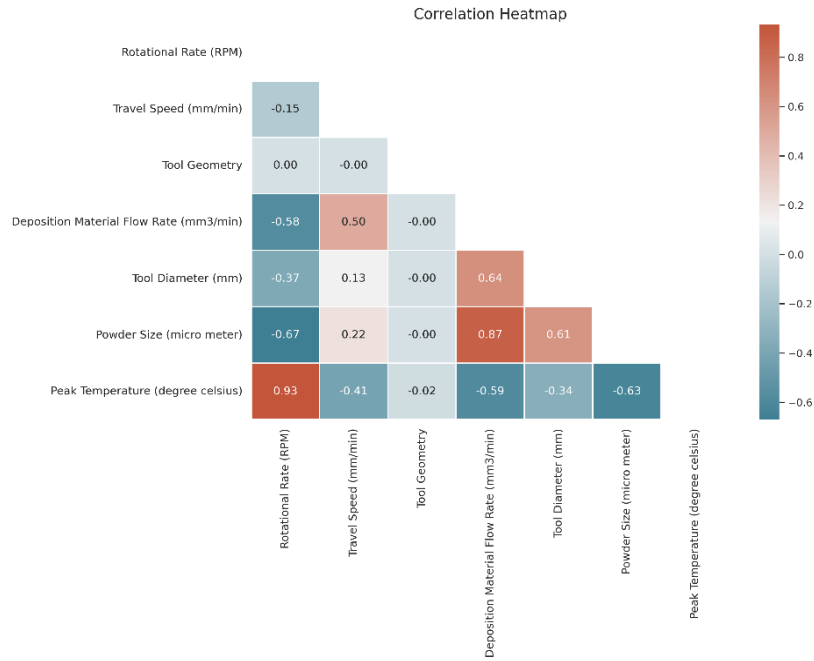

Figure 4. Obtained heat map for regression based algorithms

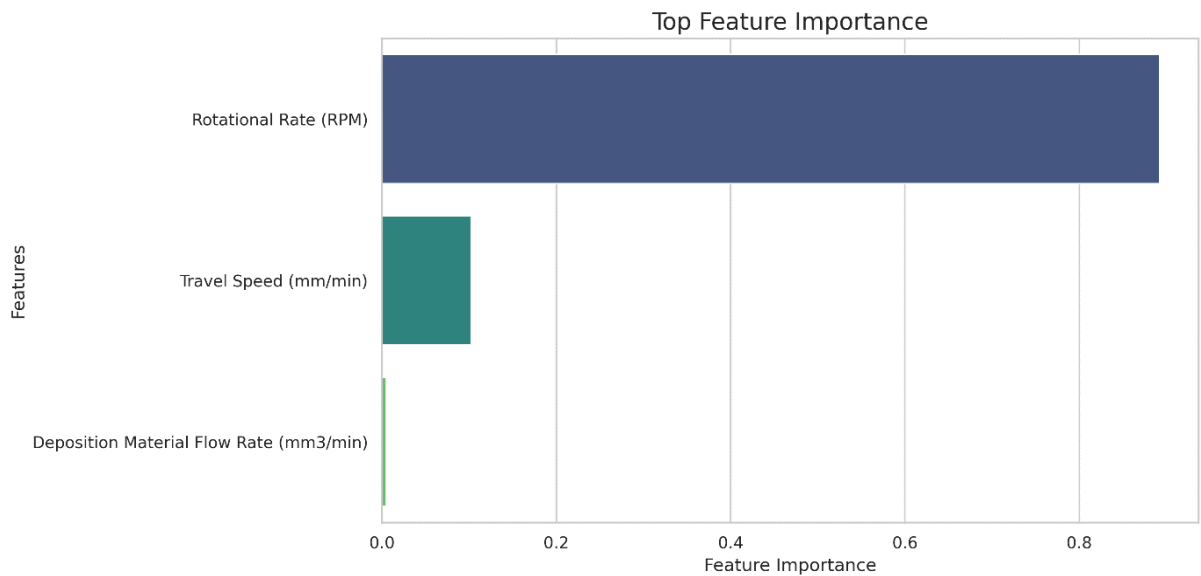

Figure 5. Feature importance plot obtained for regression based algorithms

Figure 5 shows the feature importance plot, it is observed that the Rotational rate has highest contribution towards the output parameter i.e., towards the peak temperature value. Now let's discuss about the working mechanism and results obtained from the implemented regression based algorithms to predict the peak temperature.

Support Vector Regression (SVR) is a powerful machine learning technique employed for regression tasks, particularly in cases where the relationships between input features and the output parameter are nonlinear and complex. In the context of present research, there are three input parameters: Rotational Rate (RR), Travel Speed (TS), and Deposition Material

Flow Rate (DMFR), denoted as $x_i$ = [$RR_i$, $TS_i$, $DMFR_i$]. The goal of SVR is to predict the output parameter, Peak Temperature (PT), represented as $y_i$. The primary mathematical objective of SVR can be expressed in Equation 1.

$$Minimize: \frac{1}{2}\|w\|^2 + C.\sum_{i=1}^{n}[\varepsilon_i + \varepsilon_i^*] \tag{1}$$

where 'w' corresponds to the weight vector of the hyperplane, 'C' is the regularization parameter controlling the trade-off between maximizing the margin and minimizing the error, '$\varepsilon_i$' and '$\varepsilon_i$*' are slack variables representing the degree of violation of the margin by individual data points. The hyperplane is denoted by the Equation 2.

$$f(x) = w^T . x + b \tag{2}$$

where 'f(x)' signifies the predicted Peak Temperature (PT), 'x' is the input feature vector containing Rotational Rate (RR), Travel Speed (TS), and Deposition Material Flow Rate (DMFR), and 'b' is the bias term. SVR incorporates constraints to ensure that the predicted PT values align with the actual PT values within a specified margin ($\varepsilon$) shown in Equation 3.

$$PT_i - f(x_i) \leq \varepsilon + \varepsilon_i^*, \; f(x_i) - PT_i \leq \varepsilon + \varepsilon_i^* \tag{3}$$

These constraints enforce the ability of SVR to handle errors within the margin while penalizing deviations exceeding this margin. The quantification of these deviations is performed using the epsilon-insensitive loss function shown in Equation 4.

$$L(\varepsilon, \varepsilon^*) = \max(0, |PT_i - f(x_i)| - \varepsilon - \varepsilon_i^*) \tag{4}$$

This loss function incentivizes the SVR model to minimize errors while permitting those within the specified margin. The optimization task associated with SVR entails determining the optimal hyperplane coefficients 'w' and 'b' while respecting the constraints and minimizing the loss.

When it comes to handling regression issues, Decision Tree Regression is a reliable and understandable machine learning technique. When the connections between the input characteristics and the output variable are intricate and nonlinear, this approach performs especially well. The core mathematical idea of Decision Tree Regression is repeatedly dividing the dataset into subsets depending on the input features while striving to maximize a particular criterion. The forecast for a new input data point is created by averaging the training data points' output values in the leaf node that the new data point belongs to. The algorithm specifically chooses, at each node of the tree, the feature that yields the best split, often by reducing the mean squared error (MSE) shown in Equation 5.

$$MSE = \sum \frac{(y_i - \bar{y})^2}{n} \tag{5}$$

where '$y_i$' is the actual output value, '$\bar{y}$' is the mean of the output values in the current node, and '$n$' is the number of data points in the node. The feature and threshold that minimize the MSE are chosen to make the split.

Decision Tree Regression repeats this process, adding additional nodes and splits until a stopping requirement, such as a maximum tree depth or a minimum amount of data points per leaf, is met. This produces a tree structure that maps input information to expected output values shown in Figure 6. The decision tree has a maximum depth of 3 levels and uses 'Rotational Rate (RPM)' as the feature for the root node split. This indicates that Rotational Rate is the most important feature for predicting the target variable 'Peak Temperature (degree celsius)'. The tree uses Rotational Rate as the root node split, suggesting it is the most important feature. For lower RPM ≤ 1350, it further splits on Travel Speed. This shows both RPM and Travel Speed help segment the data into regions with different temperature means.

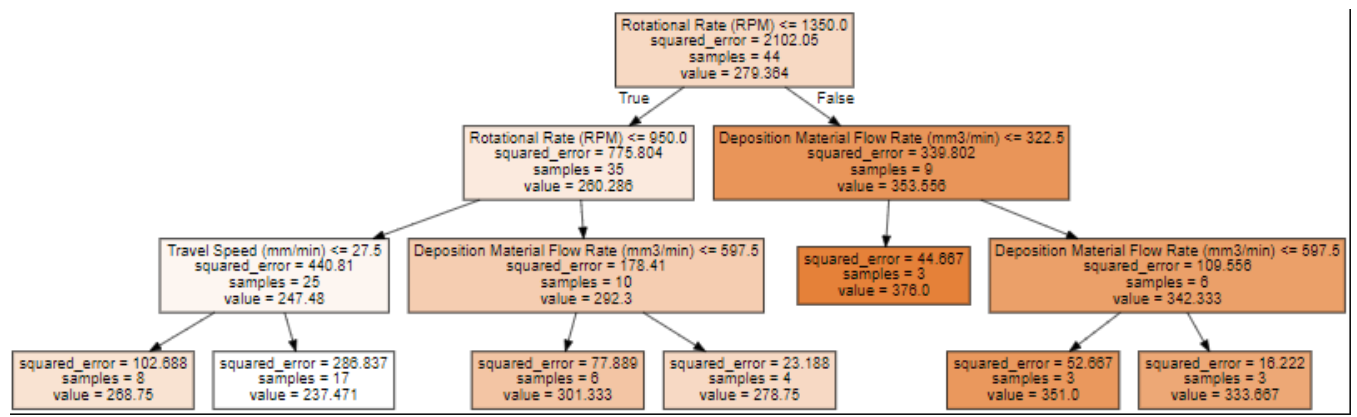

Figure 6. Decision Tree plot obtained in the present work

The Random Forest algorithm generates a set of decision trees in two steps: bootstrapped sampling and feature randomness. First, it creates numerous bootstrapped datasets by randomly selecting sections of the training data with replacement. Then, for each bootstrapped dataset, it builds a decision tree with only a random subset of the features considered at each node. The final prediction for a new input data point is derived by combining all of the individual tree forecasts. This is typically accomplished by averaging the anticipated values for regression jobs. The Random Forest ensemble reduces overfitting in individual decision trees and increases overall forecast accuracy. It is ideal for capturing complex and nonlinear interactions between input parameters and output parameters. The capacity to reduce variation while keeping low bias is critical to its effectiveness. Random Forest Regression is used in present study to estimate Peak Temperature (PT) based on the input parameters Rotational Rate (RR), Travel Speed (TS), and Deposition Material Flow Rate (DMFR). The ensemble technique improves model generalization by providing robustness against noisy data. The Random Forest ensemble is a weighted average of individual decision trees as shown in Equation 6.

$$f(x) = \frac{1}{N}\sum_{i=1}^{N} f_i(x) \qquad (6)$$

where 'f(x)' represents the final prediction, 'N' is the number of trees in the ensemble, and '$f_i(x)$' is the prediction of the i[th] decision tree.

XGBoost (Extreme Gradient Boosting) is a cutting-edge ensemble learning algorithm that solves regression problems with amazing accuracy and efficiency. This strong technique is based on gradient boosting principles, and it successively combines numerous weak predictive models (usually decision trees) into a robust and highly predictive ensemble. The core idea underlying XGBoost is to train decision trees iteratively to rectify errors generated by the ensemble of previously trained trees. This is accomplished by the minimization of a specific loss function that quantifies the difference between expected and actual output values. To control model complexity, the approach combines regularization techniques, as well as a sophisticated approximation strategy to speed up the optimization process. Furthermore, XGBoost includes features such as gradient boosting and boosting, a sophisticated ensemble strategy that iteratively updates the model using a weighted sum of the weak learners. XGBoost optimizes the objective function shown in Equation 7.

$$Obj = \sum_{i=1}^{n} L(y_i, \hat{y}_i) + \sum_{k=1}^{K} \Omega(f_k) \tag{7}$$

where 'Obj' is the overall objective, 'n' is the number of data points, '$L(y_i, \hat{y}_i)$' represents the loss function that quantifies the error for each data point, 'K' is the number of leaves in the tree, '$f_k$' represents the prediction from the k-th tree, and '$\Omega(f_k)$' is a regularization term that penalizes complex models.

CatBoost, short for Categorical Boosting, is a cutting-edge gradient boosting technique that excels at solving regression problems while handling categorical variables effectively. It is a huge step forward in the fields of ensemble learning and machine learning. CatBoost, like other gradient boosting systems, uses an ensemble of decision trees, but it adds numerous improvements that set it different. Notably, CatBoost handles categorical data natively without the need for considerable preprocessing, as it leverages techniques like as ordered boosting and oblivious trees. CatBoost also presents a novel way for dealing with overfitting using a per-leaf algorithm, effectively lowering model complexity. To train gradient-boosted ensemble models, the objective functions of XGBoost (Extreme Gradient Boosting) and CatBoost (Categorical Boosting) are similar in that they both strive to minimize a combination of a loss function and a regularization term. However, the particular formulations and specifics of the two algorithms may differ. Regularization terms are introduced by XGBoost to regulate the complexity of the individual trees in the ensemble, preventing overfitting. On the leaf scores, these regularization terms include L1 (Lasso) and L2 (Ridge) regularization. CatBoost also employs a combined loss function and regularization goal function. CatBoost, on the other hand, offers particular approaches for dealing with categorical characteristics natively and addressing overfitting.

AdaBoost, short for Adaptive Boosting, is a classification problem-solving ensemble learning technique. This approach is particularly good at integrating numerous weak classifiers to form a resilient and highly accurate ensemble classifier. AdaBoost's major strength is its capacity to adapt and provide greater weight to data points misclassified by the ensemble, allowing it to focus on the difficult situations while continuously improving classification accuracy. AdaBoost's main principle is to generate a powerful classifier by iteratively merging the predictions of weak classifiers, which are frequently simple decision stumps (decision trees with a single split). Each weak classifier is trained on the dataset, with changed weights provided to the data points in order to highlight the misclassified ones. A weighted majority vote of these weak classifiers produces the final prediction. The weights of the ensemble's weak classifiers are determined by their accuracy, with higher-accuracy classifiers having more effect. AdaBoost basically minimizes the exponential loss function shown in Equation 8.

$$L(f) = \sum_{i=1}^{n} e^{-y_i \cdot f(x_i)} \qquad (8)$$

where 'L(f)' is the loss function, 'n' is the number of data points, '$y_i$' is the true label (+1 or -1), '$f(x_i)$' is the prediction of the classifier, and 'e' is the base of the natural logarithm. AdaBoost aims to find the weak classifiers '$f_i(x)$' and their corresponding weights '$\alpha_i$' that minimize this loss function.

The Extra Trees Regressor, also known as Extremely Randomized Trees, is a regression-specific ensemble learning technique. This algorithm draws on the ideas of decision tree regression, but adds randomness and variety to improve predictive performance. The Extra Trees Regressor is distinguished by its high degree of unpredictability during the tree-building process. Extra Trees generates numerous decision trees utilizing all available characteristics and selects random feature thresholds at each node, as opposed to typical decision trees, which find optimal splits based on a selection of features. It also introduces random data subsampling for training each tree. Extra Trees decreases the danger of overfitting while increasing ensemble diversity, resulting in a more robust and generalizable regression model. The Extra Trees Regressor uses a criterion such as mean squared error (MSE) or mean absolute error (MAE) to optimize for the optimal split at each node. A fresh input data point's final prediction is derived by averaging the predictions of all the individual trees in the ensemble. While Gradient Boosting Regression is based on iteratively improving predictions by merging the outputs of numerous weak learners, often decision trees, into a strong ensemble model. A new weak learner is trained at each iteration to capture the errors or residuals of the current ensemble's predictions. These learners are intended to rectify the faults of preceding ones, reducing the overall prediction error progressively. The final prediction for a new input data point is derived by adding the predictions from all individual trees and weighting them by a learning rate. The plots for Actual peak temperature values vs predicted peak temperature values is shown in Figure 7. We can observe that most models are able to fit the general increasing trend in the data. However, Support Vector Regression shows poor fit with high variance. Decision Tree, XGBoost, CatBoost and Gradient Boosting appear to provide the best fit with less deviation from the actual values. This indicates ensemble methods like gradient boosting perform well for this regression task.

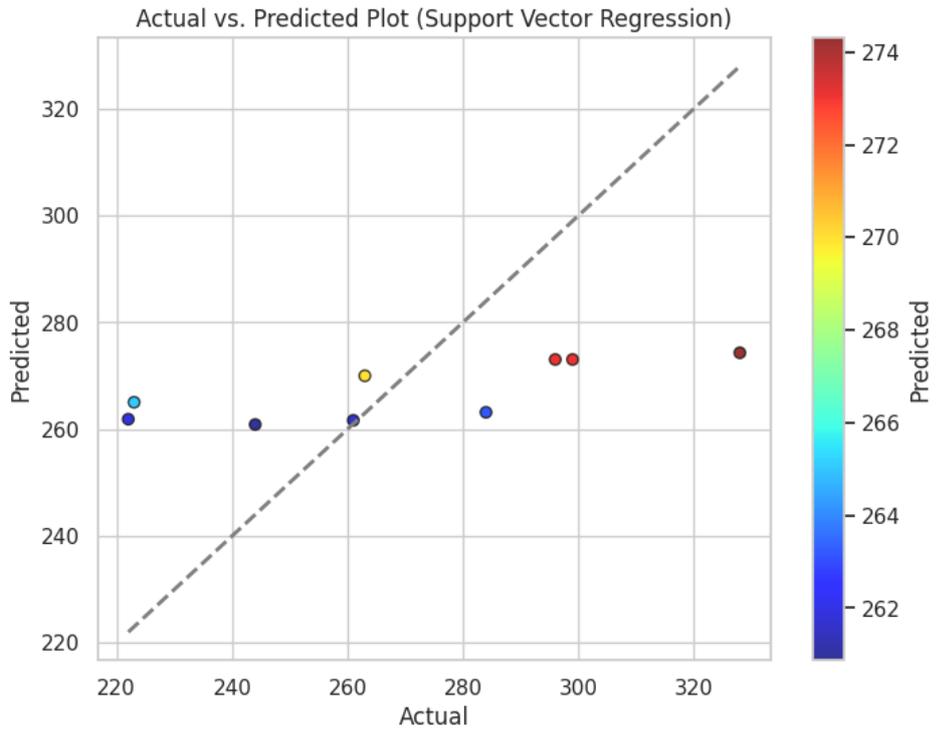

a)

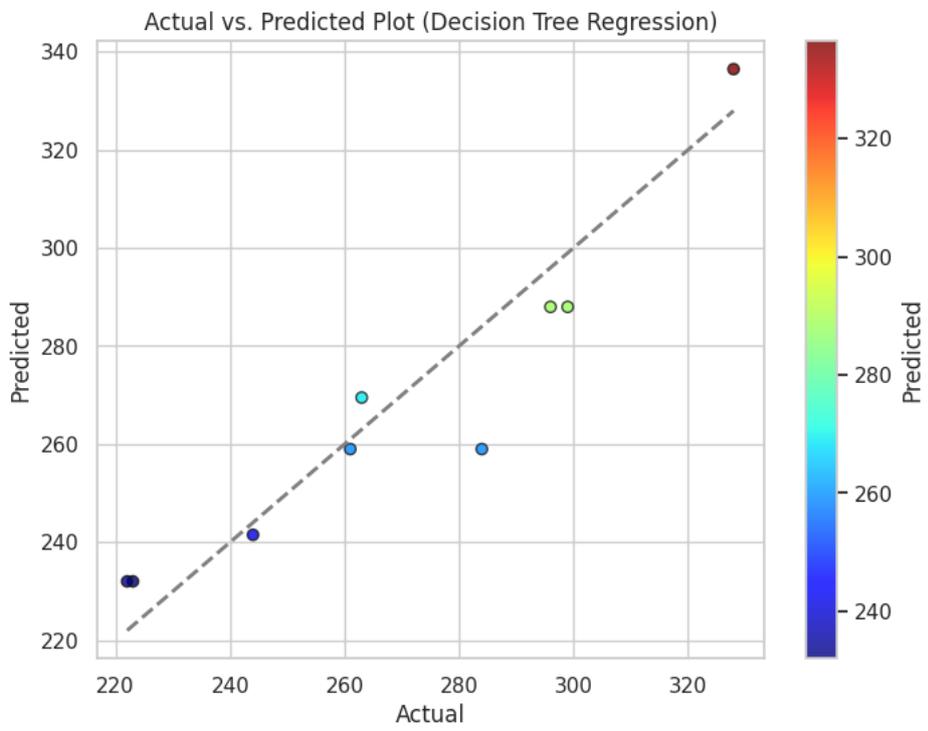

b)

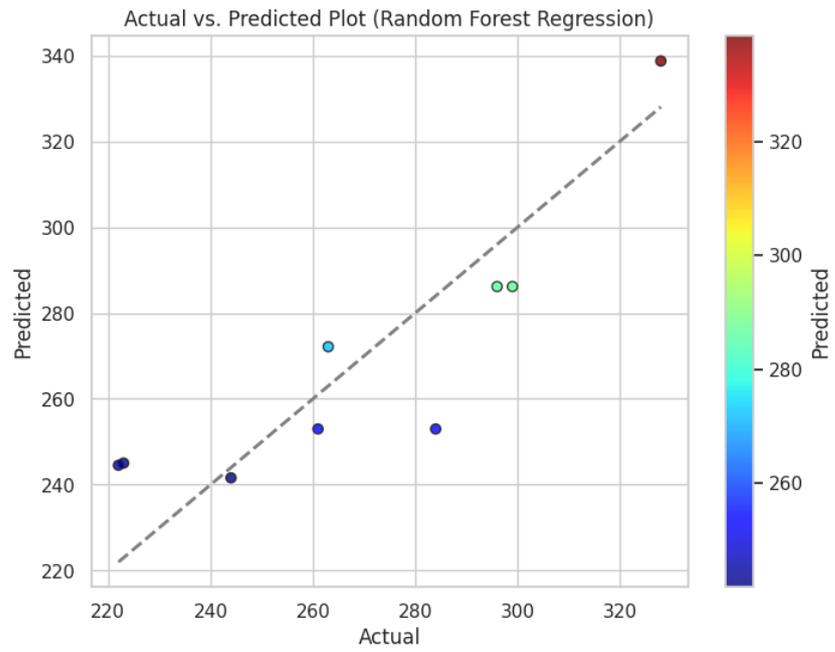

c)

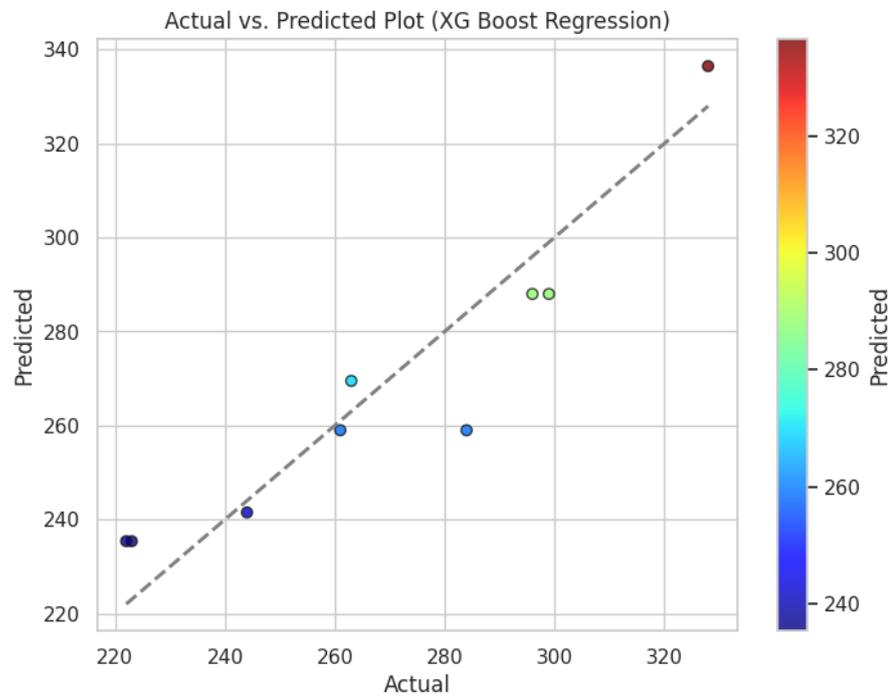

d)

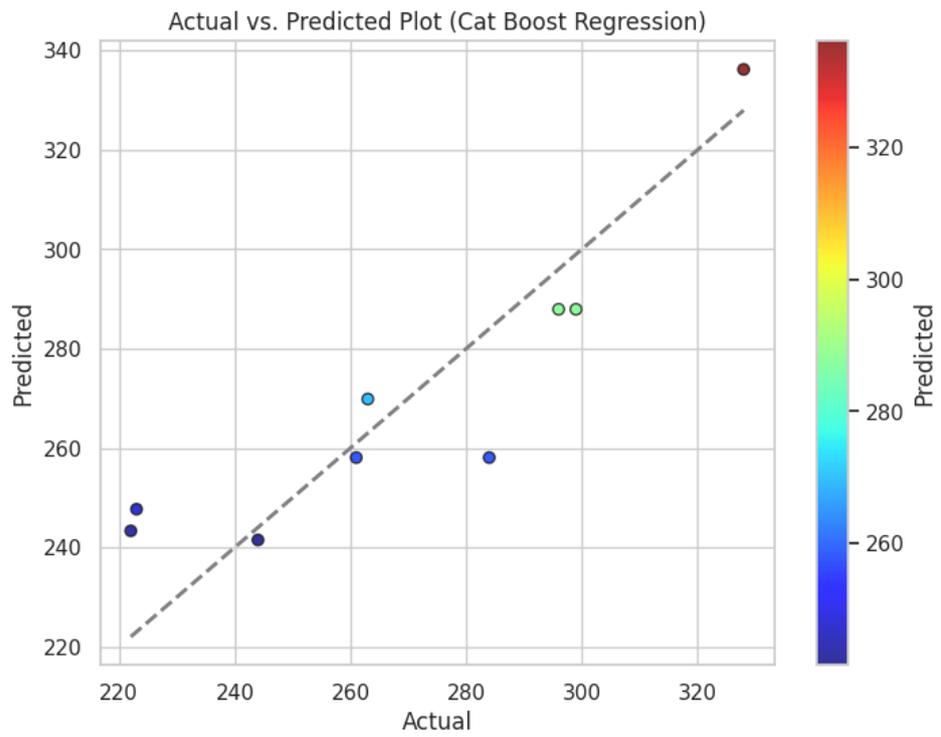

e)

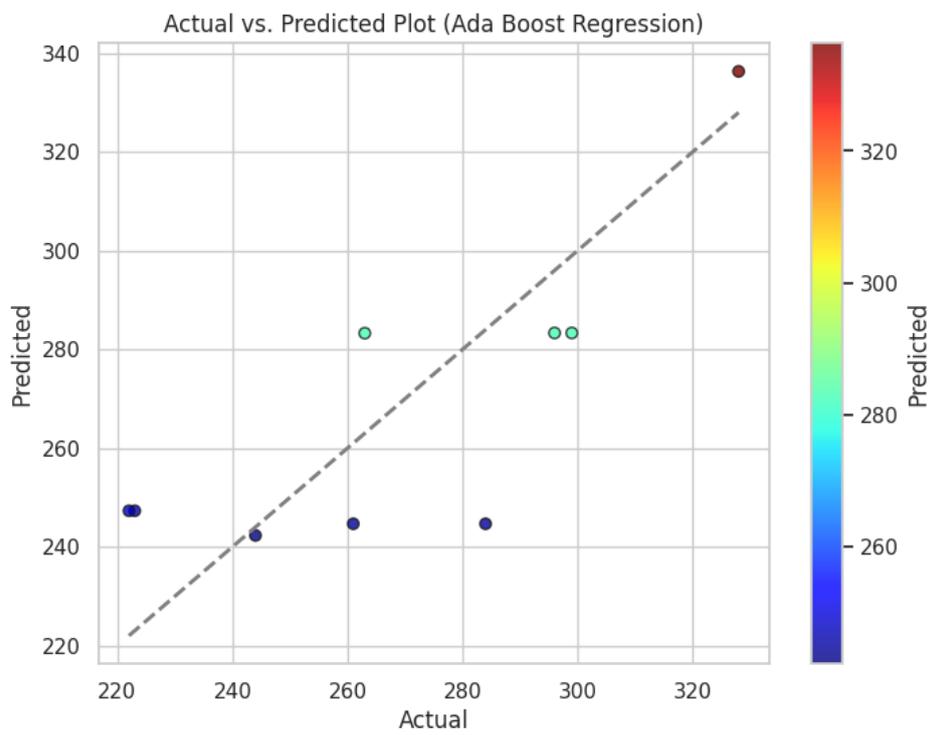

f)

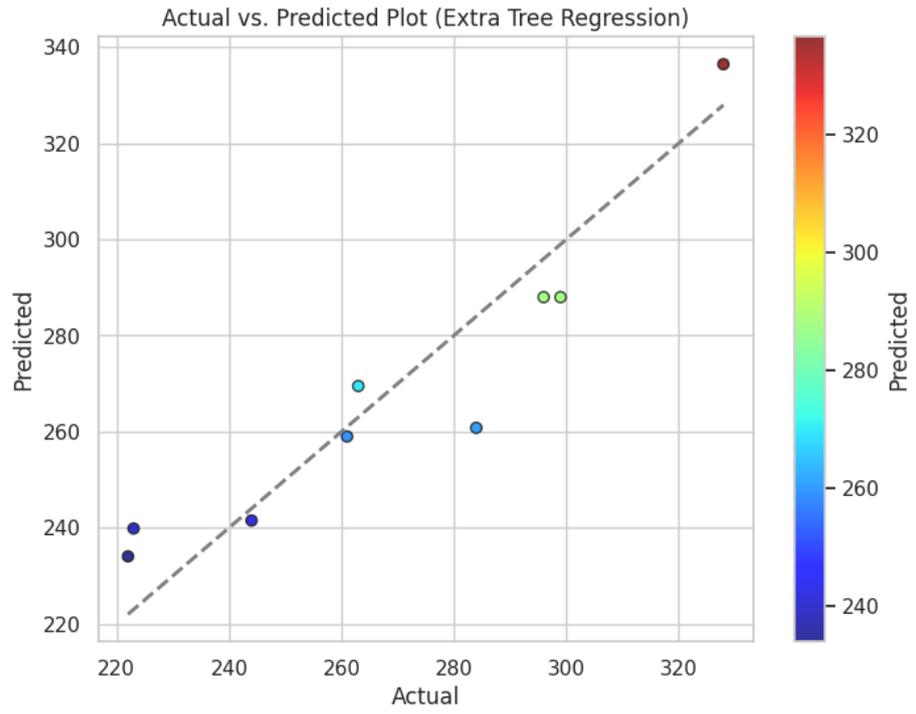

g)

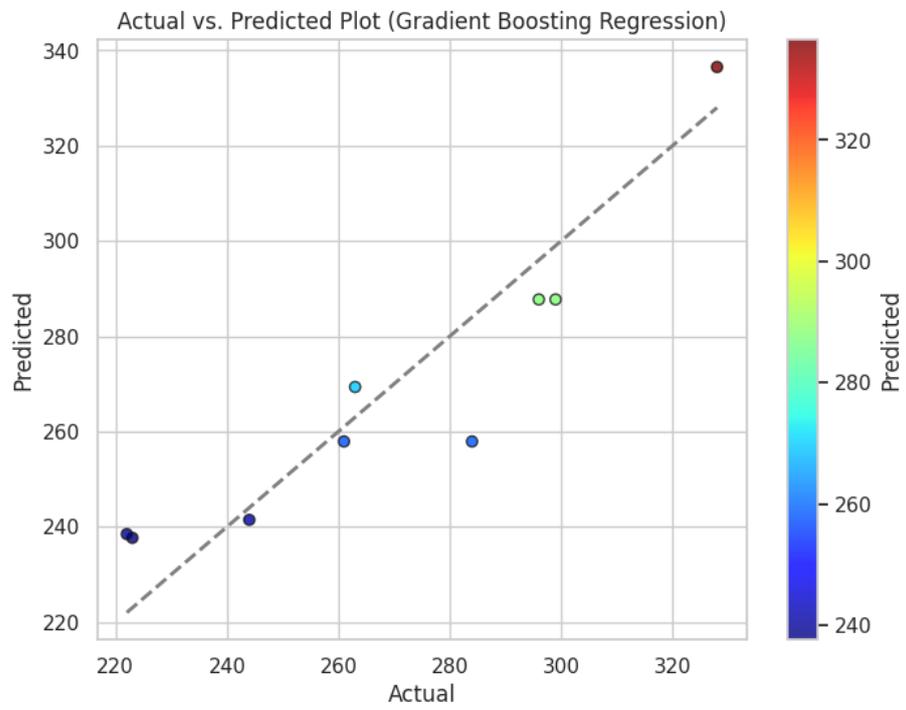

h)

Figure 7. Actual vs Predicted value plots for a) Support Vector Regression, b) Decision Tree Regression, c) Random Forest Regression, d) XGBoost Regression, e) CatBoost Regression, f) AdaBoost Regression, g) Extra Tree Regression and h) Gradient Boosting Regression algorithms

Figure 8 shows the residual plots of the implemented regression based algorithms in the present work. It is observed that the residuals (difference between actual and predicted values) are scattered randomly around 0 for most models, indicating there is no systematic pattern in the errors. However, Support Vector Regression shows a distinct funnel shape, suggesting larger errors for more extreme values.

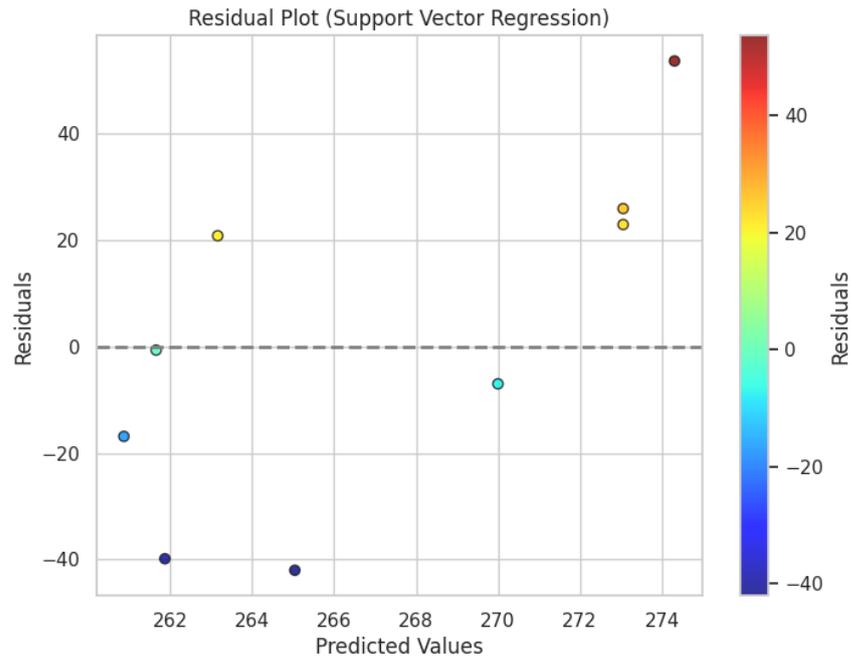

a)

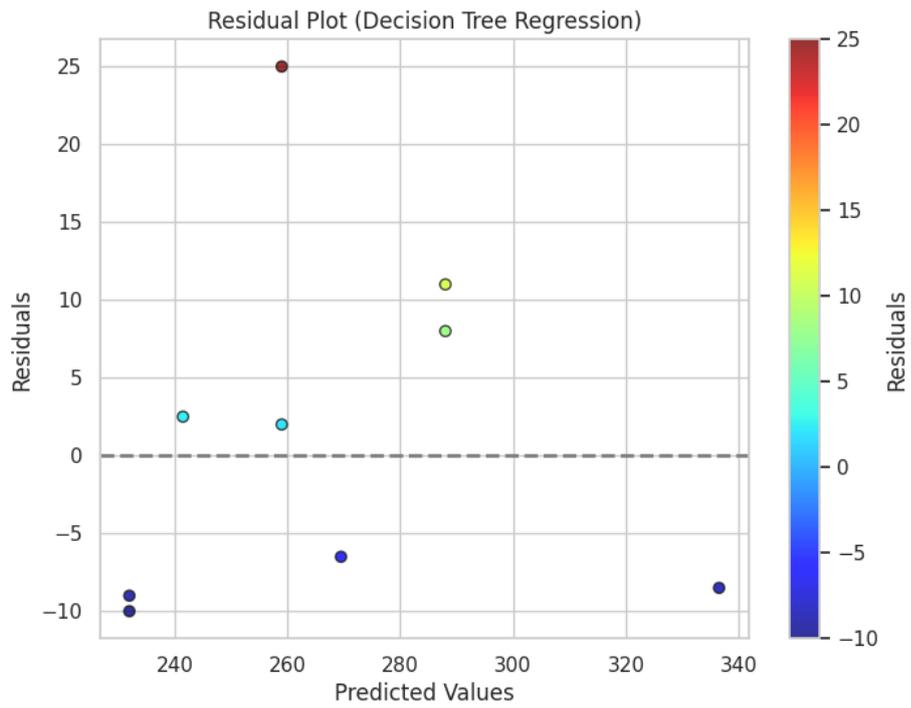

b)

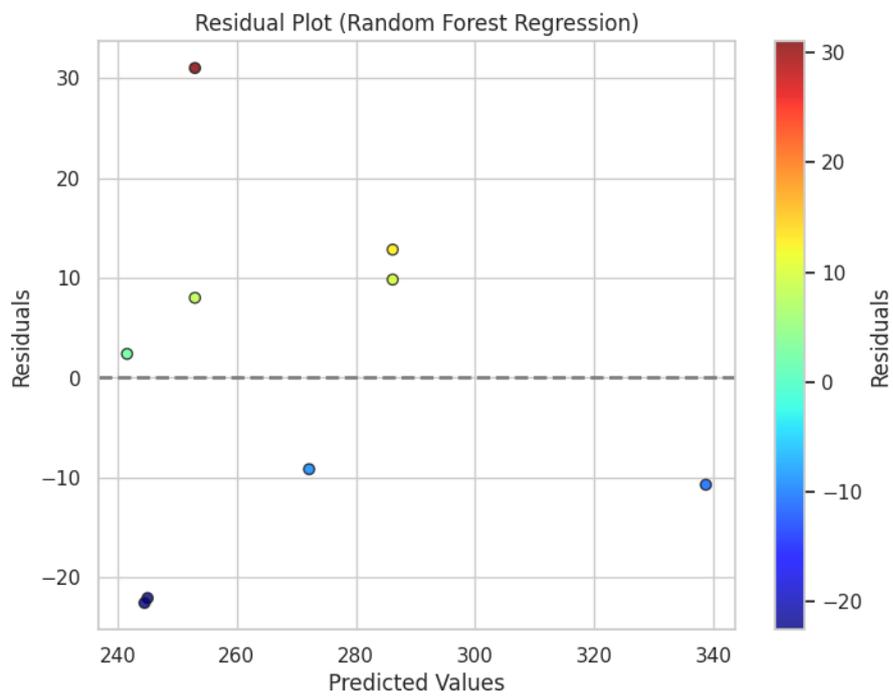

c)

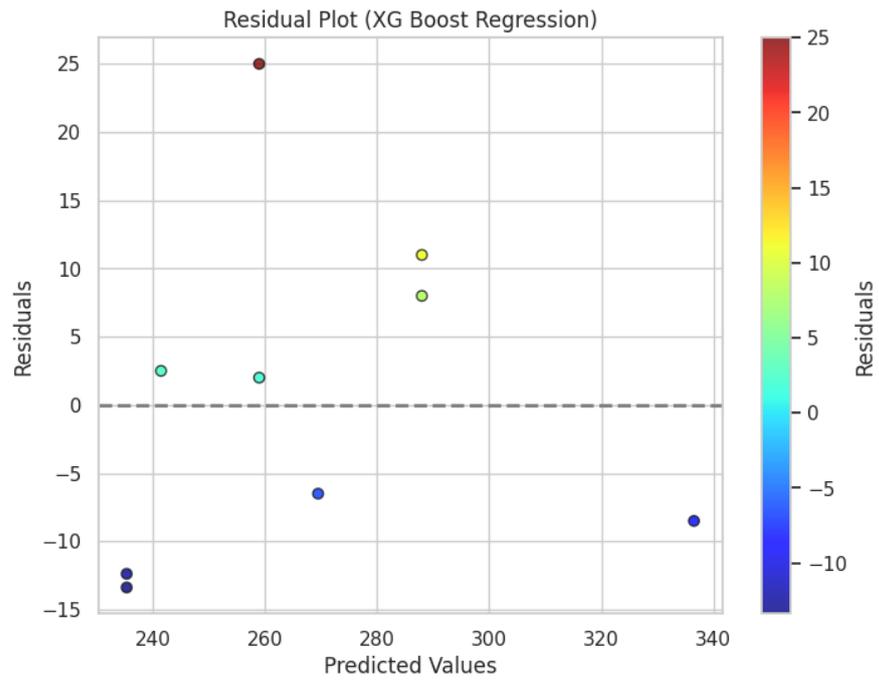

d)

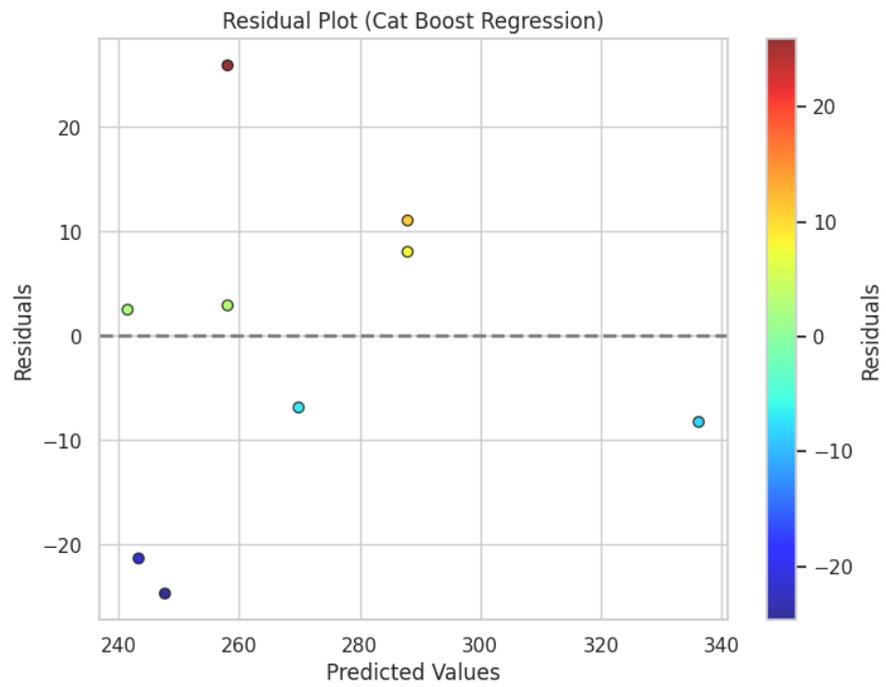

e)

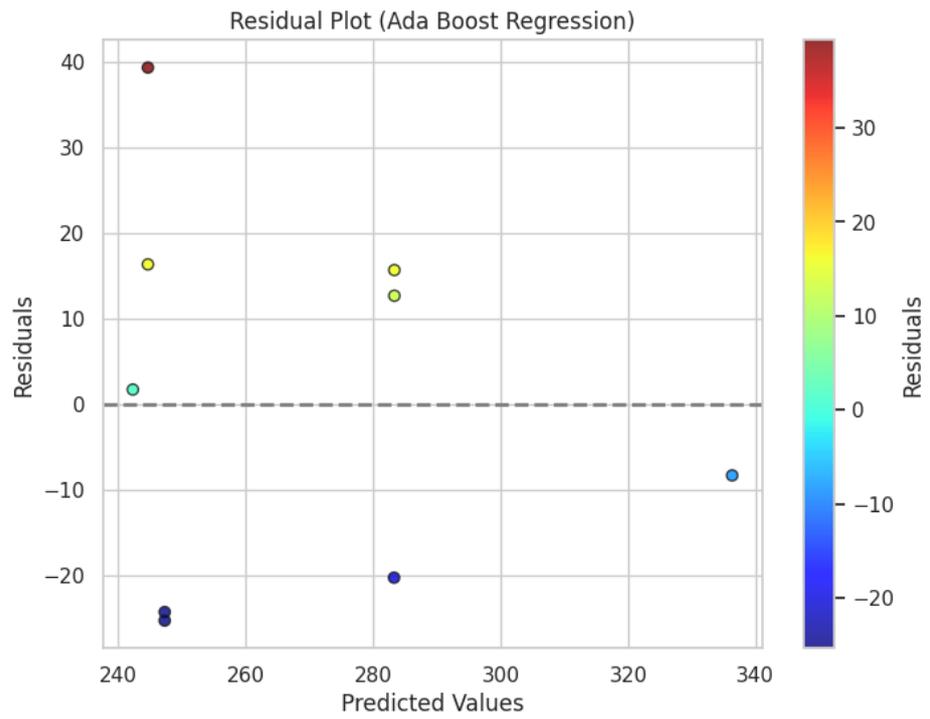

f)

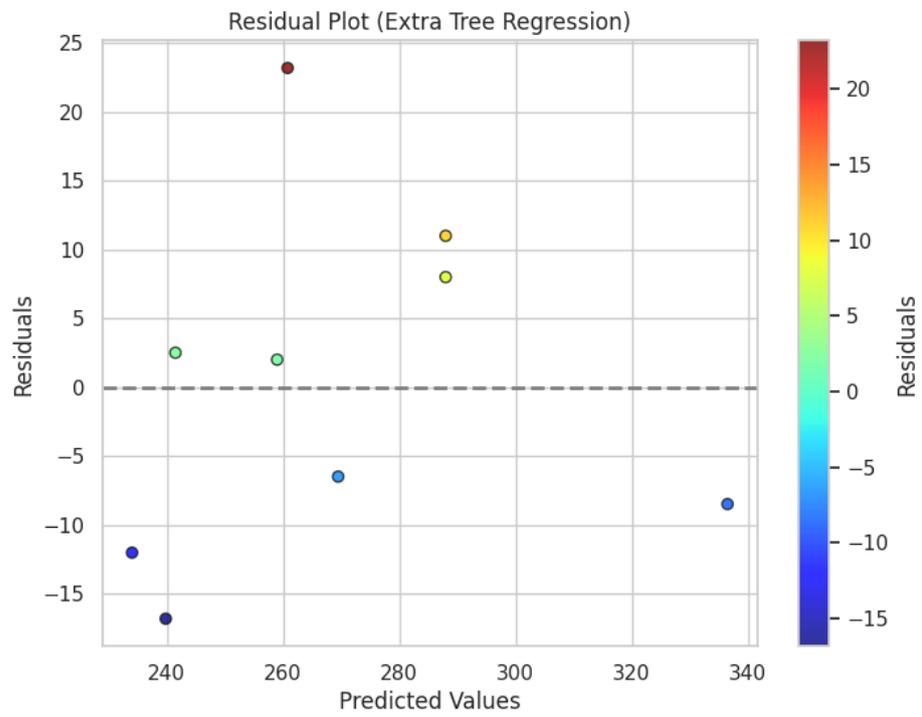

g)

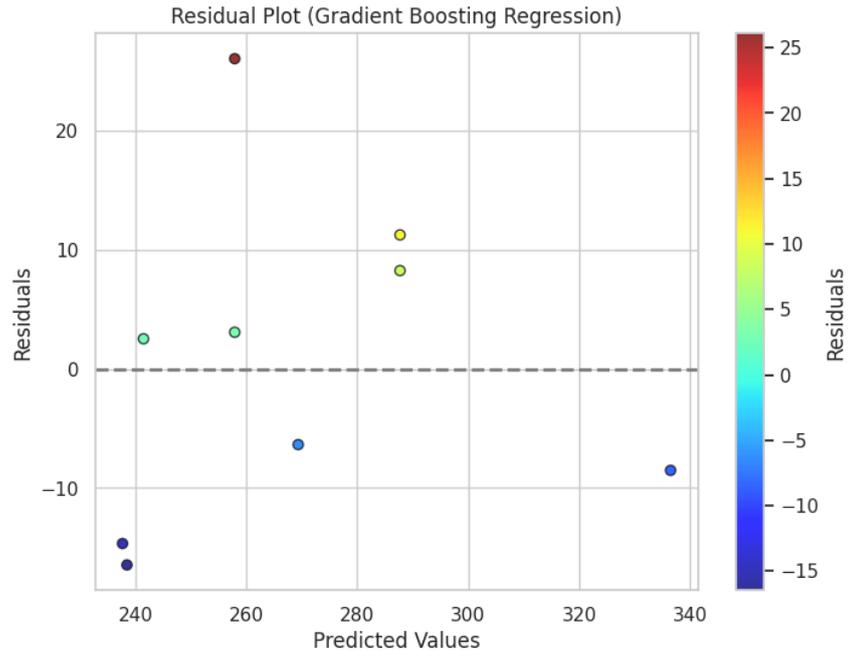

h)

Figure 8. Residual plots for a) Support Vector Regression, b) Decision Tree Regression, c) Random Forest Regression, d) XGBoost Regression, e) CatBoost Regression, f) AdaBoost Regression, g) Extra Tree Regression and h) Gradient Boosting Regression algorithms

Figure 9 displays Q-Q plots comparing the distribution of residuals to a normal distribution. For good model fit, the residuals should closely follow the normal line. We can see Decision Tree, Random Forest, XGBoost, CatBoost, and Gradient Boosting residuals align well with normal distribution. Support Vector Regression shows significant deviations, reflecting poor fit.

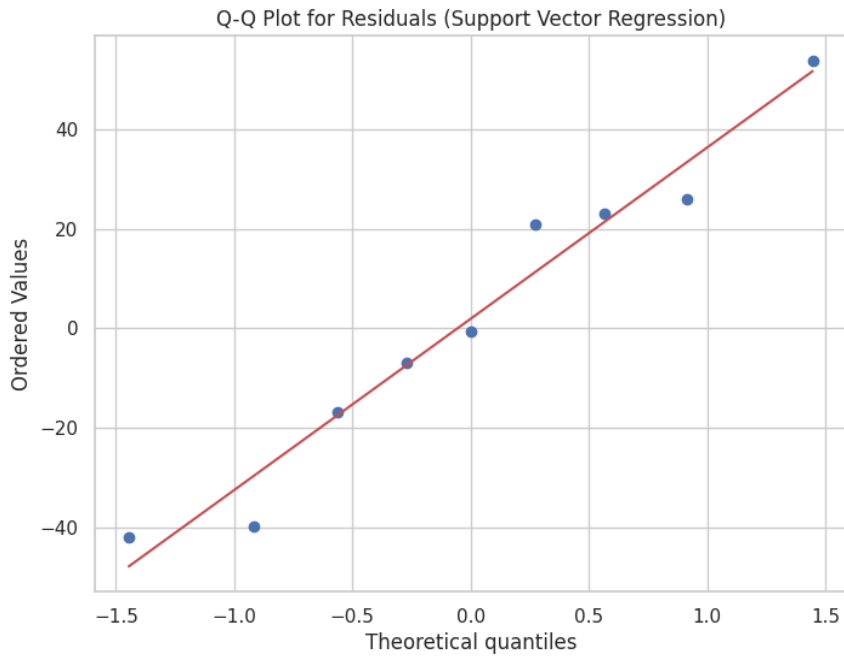

a)

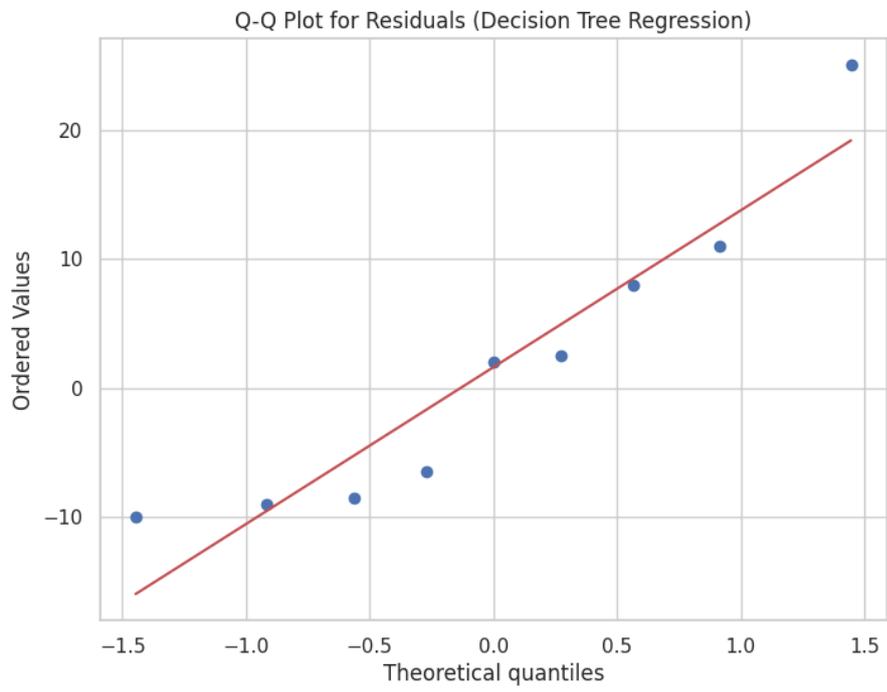

b)

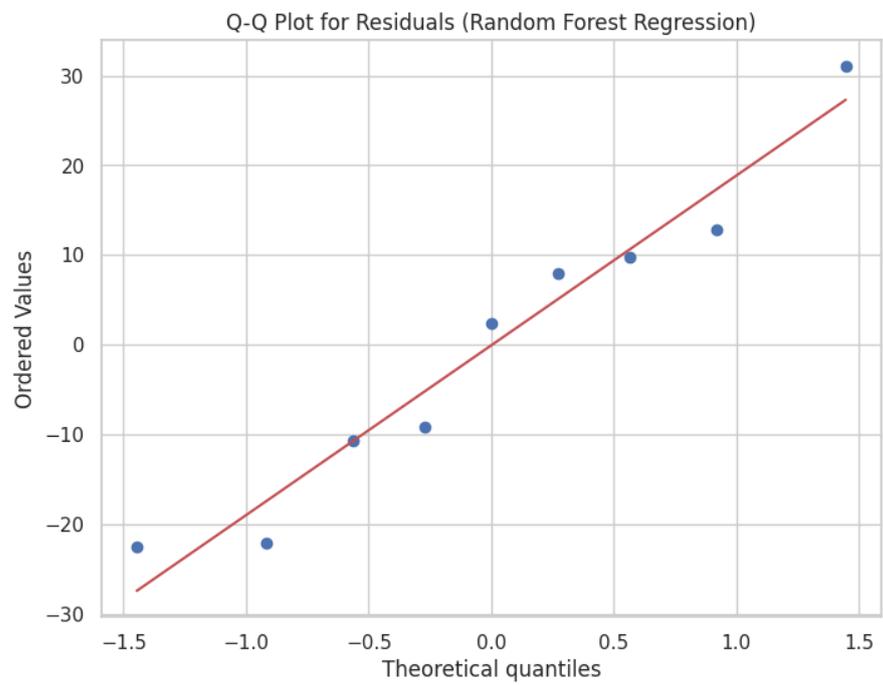

c)

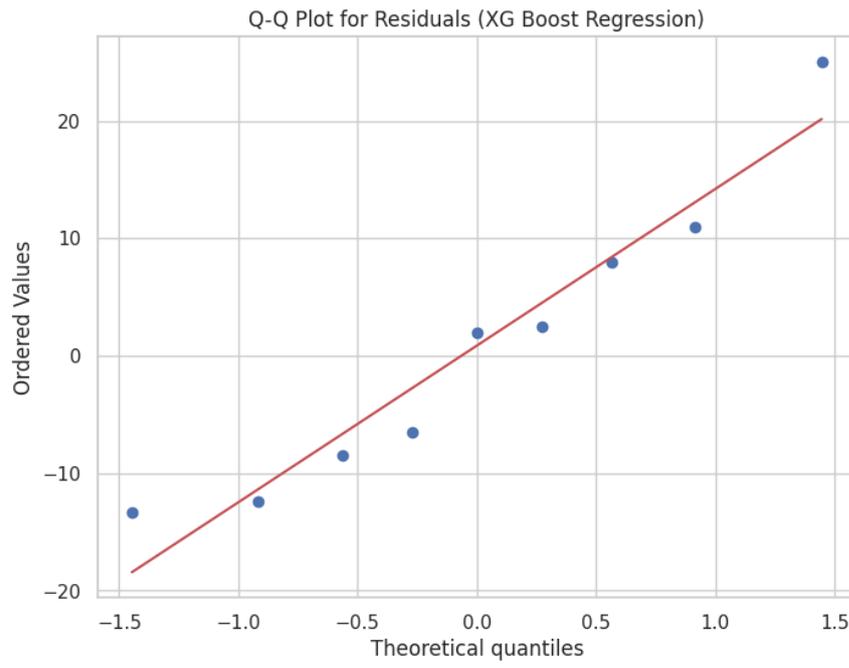

d)

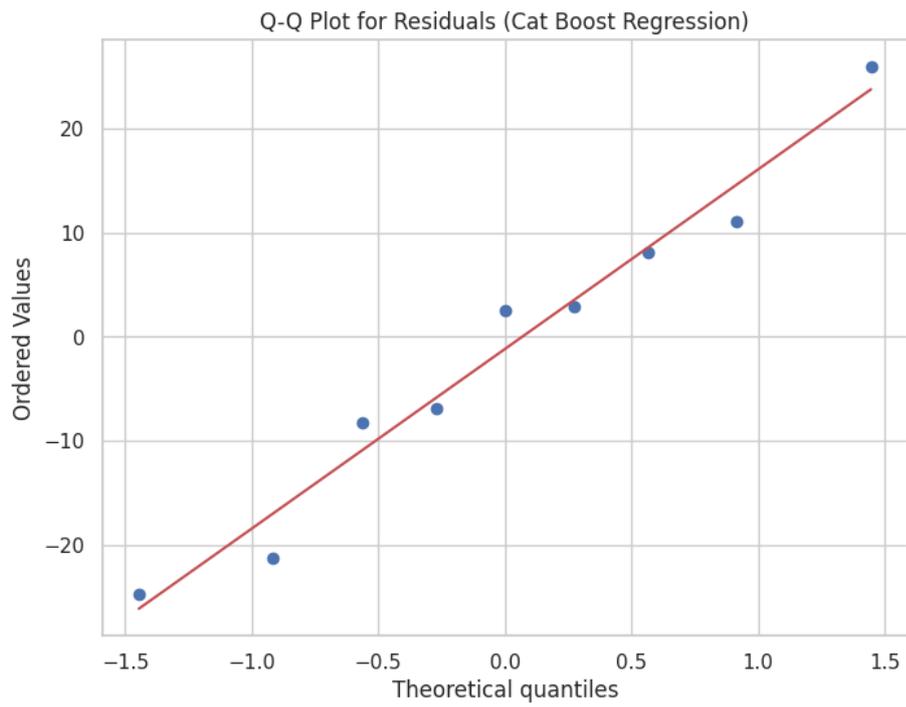

e)

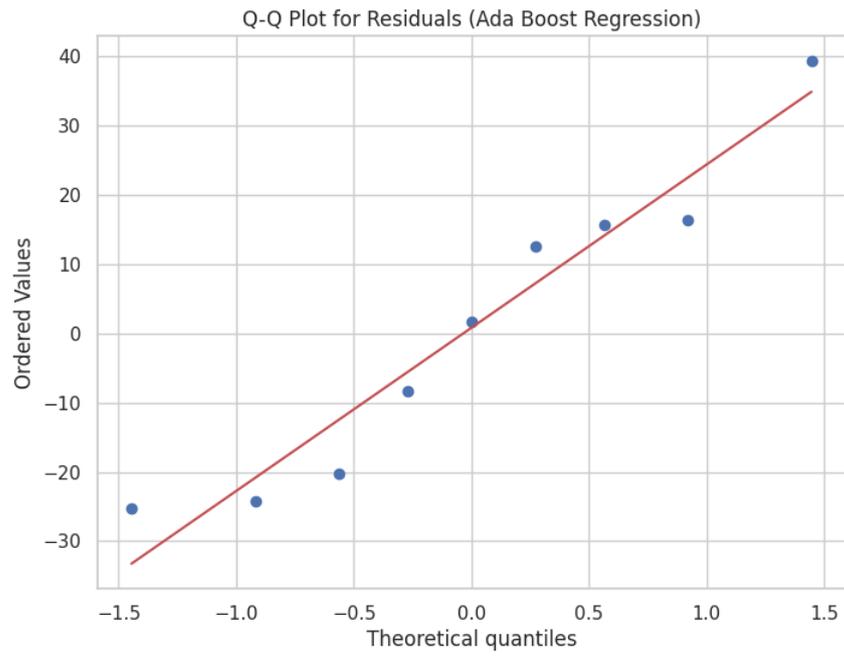

f)

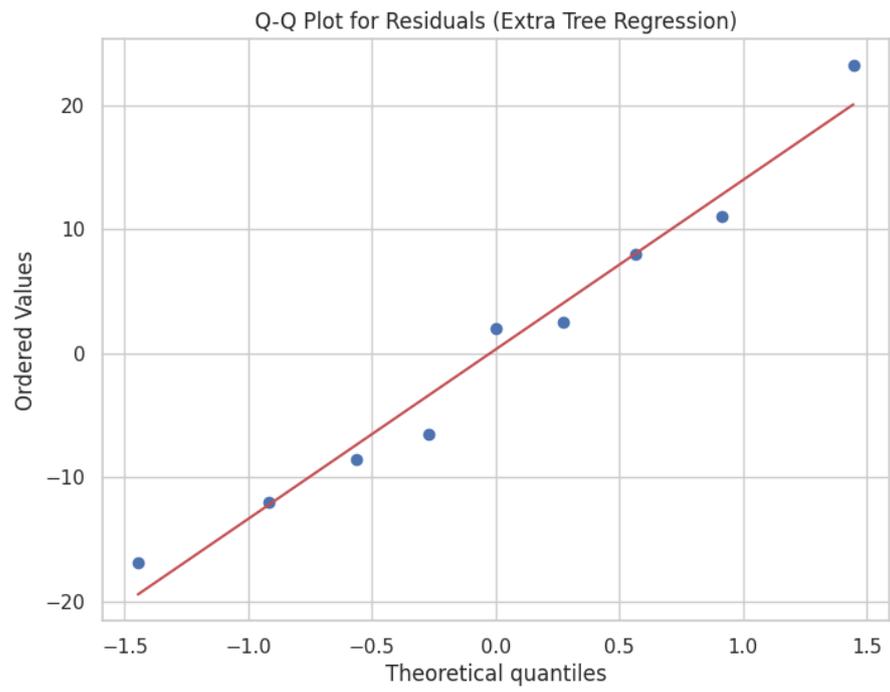

g)

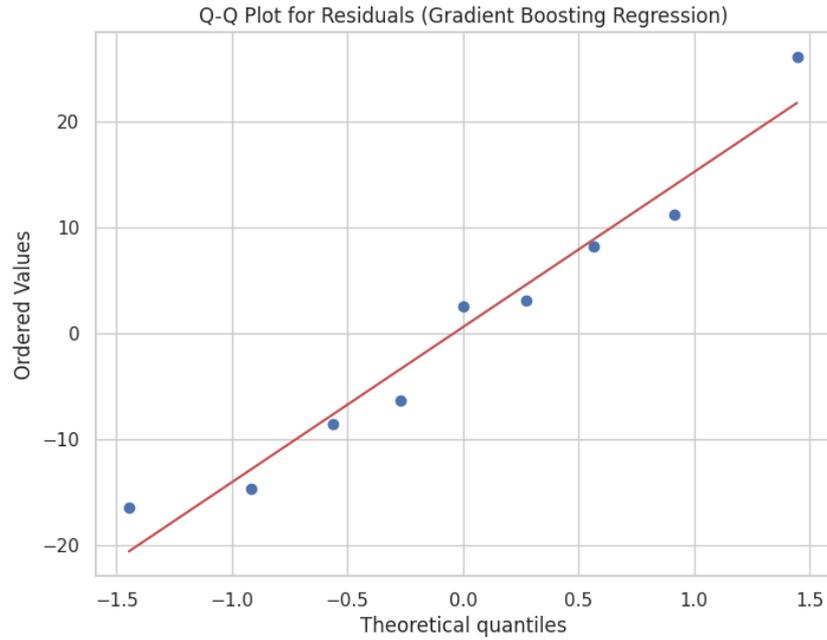

h)

Figure 9. Q-Q plots for a) Support Vector Regression, b) Decision Tree Regression, c) Random Forest Regression, d) XGBoost Regression, e) CatBoost Regression, f) AdaBoost Regression, g) Extra Tree Regression and h) Gradient Boosting Regression algorithms

Table 1. Results obtained from the used supervised machine learning regression based algorithms in the present work

| Algorithms | MSE Value | MAE Value | $R^2$ Value | Execution time in seconds |
|---|---|---|---|---|
| Support Vector Regression | 912.0466 | 25.5410 | 0.2095 | 0.0170 |
| Decision Tree Regression | 123.9722 | 9.1666 | 0.8925 | 0.0155 |
| Random Forest Regression | 276.0956 | 14.2789 | 0.7607 | 0.2282 |
| XGBoost Regression | 140.7654 | 9.9169 | 0.8780 | 0.0828 |
| CatBoost Regression | 228.0816 | 12.3990 | 0.8023 | 1.0085 |
| AdaBoost Regression | 437.1336 | 18.2132 | 0.6211 | 0.1032 |
| Extra Tree Regression | 141.7946 | 10.0629 | 0.8771 | 0.0299 |
| Gradient Boosting Rgression | 165.7845 | 10.8112 | 0.8563 | 0.1041 |

For predicting peak temperature, the ensemble methods like gradient boosting, XGBoost, random forest, and CatBoost overall outperformed simpler regression algorithms like support vector regression. This indicates ensemble techniques that combine multiple weak learners are effective for this task. Gradient boosting achieved the lowest MSE of 165.78 and a high $R^2$ of 0.8563, demonstrating strong predictive performance. XGBoost and extra trees also produced low MSE values under 200. Support vector regression performed poorly with very high MSE of 912 and low $R^2$ of 0.2095.

**4.2 Physics based Machine Learning algorithms for Peak temperature prediction**

This research also employed Physics-Informed Neural Network (PINN) to predict Peak Temperature (PT) as a continuous output parameter based on three input parameters: Rotational Rate (RR), Travel Speed (TS), and Deposition Material Flow Rate (DMFR). The research problem can be framed mathematically as the prediction of PT (y) as a function of the input features, $x_i$ = [$RR_i$, $TS_i$, $DMFR_i$]. The loss function, which combines two fundamental components i.e. Physics-Informed Loss (PINN Loss) and Data-Driven Loss, is at the heart of the PINN model.

The transport equation lies at the heart of this technique. It entails computing the derivatives of model predictions (u) with respect to time (t) and position (x). The transfer equation is written as shown in Equation 9.

$$c \cdot \frac{\partial u}{\partial t} + \frac{\partial u}{\partial x} = 0 \tag{9}$$

Where 'c' represents the advection velocity. The PINN loss term, denoted as 'pinn loss,' ensures that the derivatives of the model predictions adhere to this transport equation, enforcing the physics-based constraints.

The data-driven component quantifies the error between the model's predictions and the actual target values. It is expressed as shown in Equation 10.

$$data\ loss = \frac{1}{n}\sum_{i=1}^{n}(u_i - y_i)^2 \tag{10}$$

Where '$u_i$' represents the model's prediction for the i-th data point, '$y_i$' is the actual target value (Peak Temperature), and 'n' is the number of data points.

The overall loss function, 'loss,' combines these two components as shown in Equation 11.

$$loss = pinn\ loss + data\ loss \tag{11}$$

During training, the PINN model minimizes this combined loss. Gradients are calculated using TensorFlow's GradientTape, and gradient descent is applied using the Adam optimizer.

The model learns to represent the complicated interactions between the input parameters (RR, TS, DMFR) and the output (PT) while adhering to the physics-driven limitations defined by the transport equation during the training phase. Following training, the model is used to predict on a separate testing dataset, and assessment measures such as Mean Squared Error (MSE) and Mean Absolute Error (MAE) are calculated to evaluate its predictive accuracy.

Now let's discuss about the Physics based Machine Learning model based on the wave equation in terms of predicting the output parameter, Peak Temperature (PT). In the context of predicting PT, the fundamental physics is represented by the wave equation shown in Equation 12. This equation captures the physics of wave behavior, indicating how the second derivatives of 'u' in both time and position are related.

$$c^2 \cdot \frac{\partial^2 u}{\partial t^2} - \frac{\partial^2 u}{\partial x^2} = 0 \tag{12}$$

Where u' represents the wave function (in this case, Peak Temperature, PT) and 'c' is the wave velocity, which characterizes how fast the wave propagates.

The physics-informed loss term, 'wave loss,' assures that the model follows the wave equation as shown in Equation 13. The term 'wave loss' quantifies how well the model meets the physics-based wave equation, encouraging it to capture the physical behavior of the wave.

$$wave\ loss = \frac{1}{n}\sum_{i=1}^{n}\left(\left(c^2 \cdot \frac{\partial^2 u}{\partial t^2} - \frac{\partial^2 u}{\partial x^2}\right)^2\right) \tag{13}$$

The data driven loss is calculated by using the Equation 10 and the overall loss is calculated by using the Equation 14.

$$loss = wave\ loss + data\ loss \tag{14}$$

This complete loss metric directs the PINN model's training process, allowing it to provide accurate Peak Temperature predictions while adhering to the physics-based wave equation and efficiently fitting the given data.

In case of the physics-based machine learning model based on heat equation, the prediction is based on the principles of the heat equation, which governs how temperature evolves over time and space. The heat equation, which describes the evolution of temperature, is defined as in Equation 15.

$$\frac{\partial u}{\partial t} - k\frac{\partial^2 u}{\partial x^2} = 0 \tag{15}$$

Where 'u' represents temperature (PT in this case) and 'k' is the heat diffusion coefficient.

The physics driven loss is calculated by using Equation 16 and the Equation 10 is used for calculating the data driven loss.

Now let's discuss about the last Physics-based machine learning model used in the present work which is based on the Schrödinger equation that describes the behavior of a quantum wave function ('ψ') and is defined as Equation 16.

$$\widehat{H}\psi = i\hbar\frac{\partial \psi}{\partial t} \tag{16}$$

Where 'ψ' represents the wave function, which is the model's output, $\hat{H}$ is the Hamiltonian operator, which characterizes the energy of the quantum system, 'i' is the imaginary unit and '$\hbar$' is the reduced Planck's constant. The physics-informed loss term,'schrodinger loss,' verifies that the model follows the Schrödinger equation as shown in Equation 17.

$$Schrödinger\ loss = \frac{1}{n}\sum_{i=1}^{n}\left(\left(\hat{H}\psi - i\hbar\frac{\partial\psi}{\partial t}\right)^2\right) \qquad (17)$$

The surface plot and contour plot of the used physics based machine learning model is shown in Figure 10 and Figure 11. Table 2 shows the obtained metric features for physics based machine learning algorithms.

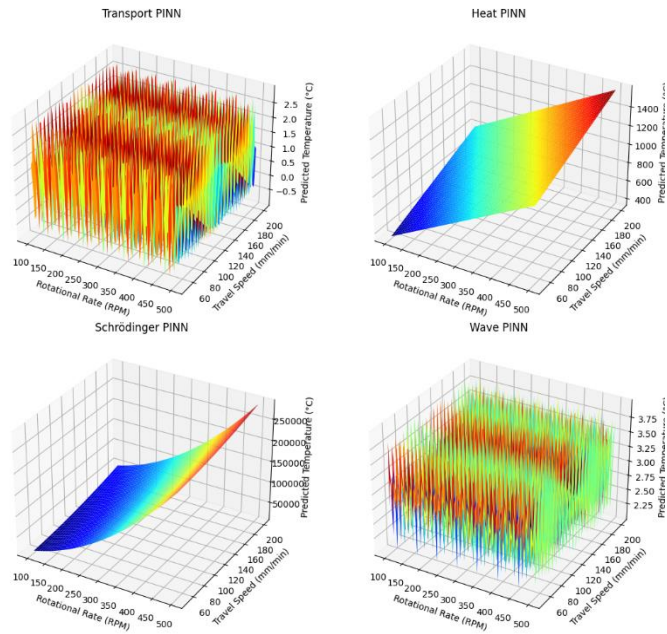

Figure 10. Surface plots of the used Physics based machine learning models in the present work

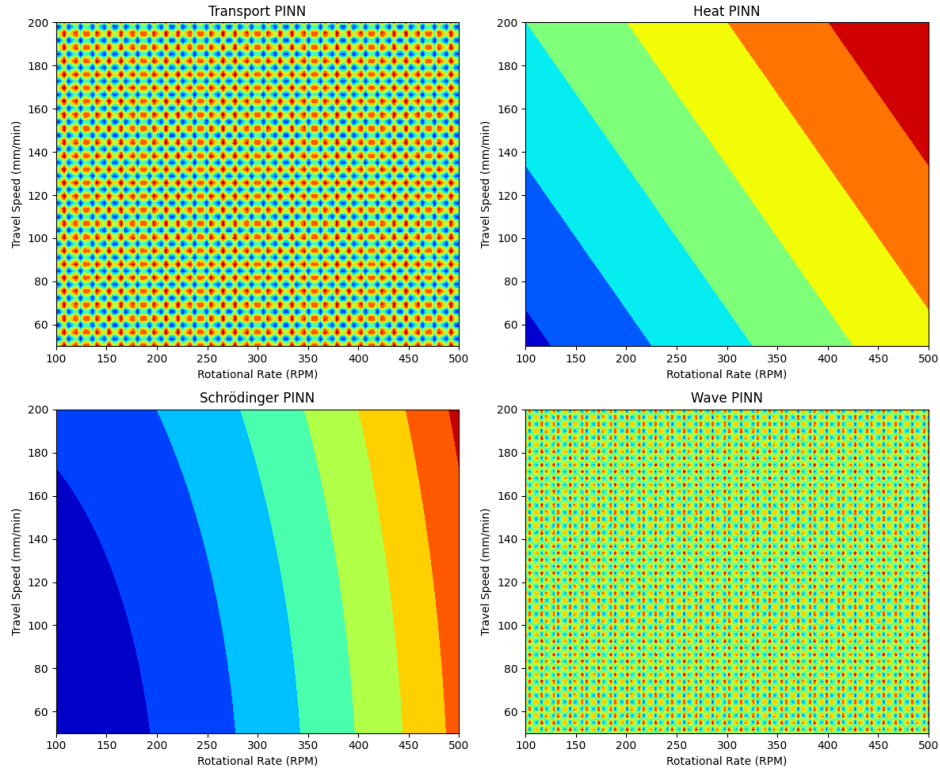

Figure 11. Contour Plot of Physics based machine learning models used in the present work

It is observed that the response surfaces are relatively smooth, with no sharp discontinuities or irregularities. This indicates the physics-informed neural networks are modeling the system smoothly and capturing coherent underlying dynamics. The peak temperature generally increases as rotational rate and travel speed increase, reflecting proper modeling of the process physics. There are some subtle differences between the models. For example, the heat equation model predicts slightly higher temperatures compared to the transport equation model under the same conditions. The contour plots also show smooth and evenly spaced contours, reaffirming continuous and well-behaved modeling by the physics-informed NNs. The contour lines are oriented diagonally, visualizing the coupled effect of rotational rate and travel speed on peak temperature.

Table 2. Results obtained from Physics based machine learning models

| Algorithms | RMSE Value | MAE value | Execution Time in seconds |
|---|---|---|---|
| Transport equation-based PINN model | 64.6096 | 57.2337 | 83.5613 |
| Wave equation-based PINN model | 66.3619 | 55.9323 | 84.6350 |
| Heat equation-based PINN model | 65.2640 | 58.8696 | 84.7443 |
| *Schrödinger* based PINN model | 71.0866 | 61.4765 | 89.9683 |

The transport equation PINN model had the lowest RMSE of 64.6 and MAE of 57.2 among the physics-informed approaches. Heat equation PINN also produced low errors with RMSE of 65.2 and MAE of 58.9. This demonstrates including physics knowledge about the underlying relationships dramatically improves predictive performance compared to pure data-driven supervised learning. Executing times for the PINN models ranged 83-90 seconds, comparable to the slower supervised algorithms like CatBoost. However, their accuracy gains show the value of added physics-based constraints.

**4.3. Supervised Machine Learning Classification based algorithms used for deposition quality prediction**

Figure 12 shows the correlation heatmap for predicting deposition quality. We can see Deposition material rate has the strongest correlation with the target variable. Rotational Rate has weaker positive correlation. This suggests deposition material rate is the most important predictor of deposition quality.

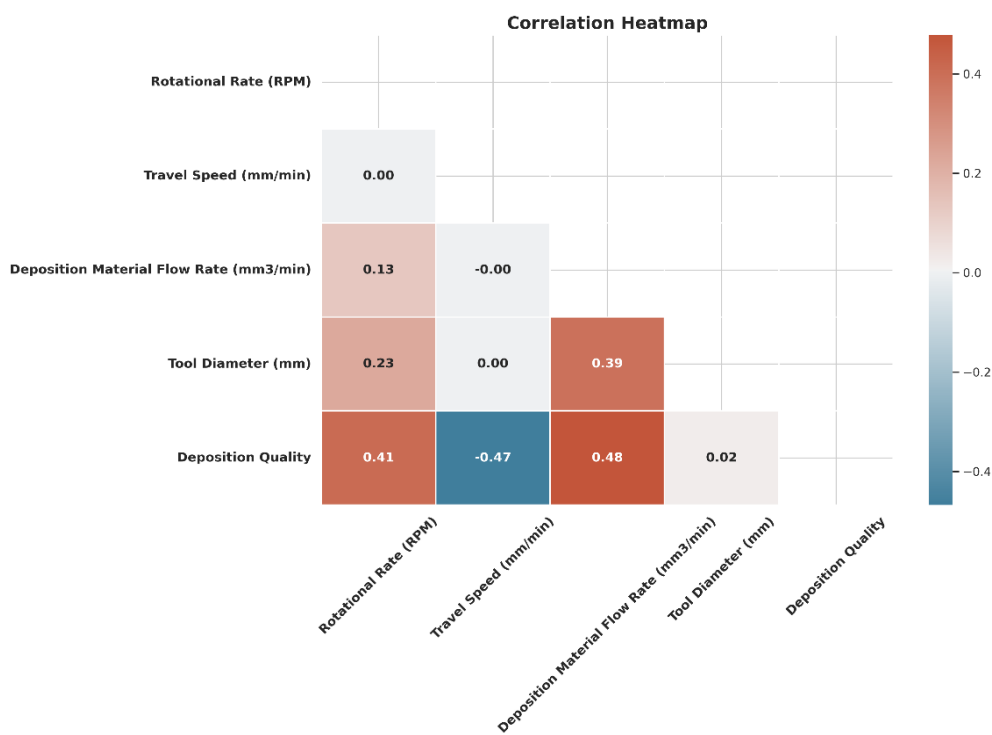

Figure 12. Correlation heatmap obtained to classify the deposition quality in the present work

Figure 13 shows the feature importance plot. It is observed that the Travel Speed (mm/min) has the highest contribution towards the deposition quality while tool diameter (mm) has no contribution towards the deposition quality. So, before training the models the tool diameter feature is dropped.

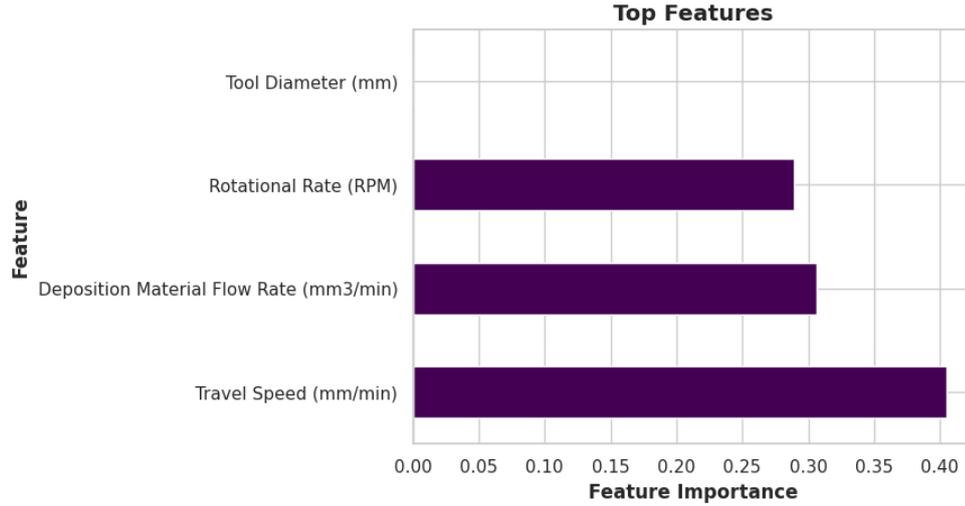

Figure 13. Feature importance plot obtained in the present work

Logistic Regression is a statistical method used for binary classification tasks in which the goal is to predict one of two possible classes (in present research work, "0" for poor deposition and "1" for good deposition) based on one or more input features (Rotational rate, Travel speed, and Deposition material flow rate). Logistic regression begins by solving a linear equation to calculate a linear combination of the input features. With three input features, it can be described in the present research work as shown in Equation 18.

$$z = \beta_0 + \beta_1 . Rotational\ Rate + \beta_2 . Travel\ Speed + \beta_3 . Deposition\ Material\ Flow\ Rate \quad (18)$$

Where $z$ is the linear combination of the input features, $\beta_0$ is the intercept term, $\beta_1$, $\beta_2$, and $\beta_3$ are coefficients associated with each input feature.

The linear combination z is then sent via a logistic function (also known as a sigmoid function), which converts it to a number between 0 and 1. The logistic function is described as in Equation 19.

$$\sigma(z) = \frac{1}{1+e^{-z}} \quad (19)$$

Where $\sigma(z)$ represents the probability that the target variable belongs to class "1" (good deposition). Because the logistic function assures that the anticipated values lie inside the range [0, 1], it is appropriate for modeling probabilities. Following the calculation of $\sigma(z)$, a threshold (typically 0.5) is used to establish the final anticipated class label. If $\sigma(z)$ is greater than or equal to 0.5, the occurrence is categorized as "1" (excellent deposition), otherwise as "0" (poor deposition).

In K-Nearest Neighbours, the prediction step comprises determining the distance between the new data point (query point) and all data points in the training dataset. The Euclidean distance, which may be represented as Equation 20, is widely employed for this purpose.

$$Eucledian\ Distance\ (X, X_i) = \sqrt{\sum_{j=1}^{n}(X - X_i)^2} \tag{20}$$

Where $X$ is the feature vector of the query point, $X_i$ is the feature vector of the *i*-th data point in the training dataset and *n* is the number of features. When using K-NN for binary classification (predicting deposition quality as "0" for bad and "1" for good), the algorithm counts the number of neighbors from each class among the K nearest neighbors. It selects the class label that appears the most frequently among the query point's K neighbors as the anticipated class.

Support Vector Classification seeks a hyperplane (a decision boundary) that best separates data points from distinct classes while maximizing margin. The data points nearest to the hyperplane are known as support vectors, and they are critical in establishing the margin and generating the decision boundary. The goal is to create a hyperplane that, for all support vectors $X_i$, fulfills the following equation, as shown in Equation 21.

$$y_i(w.X_i + b) \geq 1 \tag{21}$$

Where $y_i$ is the class label of the i-th data point ($y_i = 1$ for positive class and $y_i = 0$ for the negative class), w is the weight vector (normal to the hyperplane), $X_i$ is the feature vector of the *i*-th data point and b is the bias term. Data is frequently not perfectly separated by a hyperplane in real-world circumstances. SVC offers the concept of a "soft margin" in order to allow for occasional misclassifications while still increasing margins. This is accomplished by including a parameter C that regulates the trade-off between margin maximization and misclassification tolerance.

On the other hand, the mathematical formulation of the parameter update in Stochastic Gradient Descent (SGD) is shown in Equation 22.

$$\theta = \theta - \alpha.\nabla J(\theta) \tag{22}$$

Where $\theta$ represents the model parameters, $\alpha$ is the learning rate and $\nabla J(\theta)$ is the gradient of the cost function with respect to θ. At each iteration, SGD introduces randomization by selecting random data points (or mini-batches). This randomization aids in avoiding local minima and may result in faster convergence. While SGD converges to the cost function's minimum, it may oscillate about it. It does, however, frequently arrive at a "good enough" solution faster than classical gradient descent.

The mathematical framework of Decision Tree Classification consists mostly of calculating impurity measurements shown in Equation 23 and picking the appropriate feature to partition the data as shown in Figure 14.

$$Impurity = \sum_{i=1}^{n} p(i).(1 - p(i)) \tag{23}$$

Where n is the number of classes and $p(i)$ is the proportion of data points that belongs to class i.

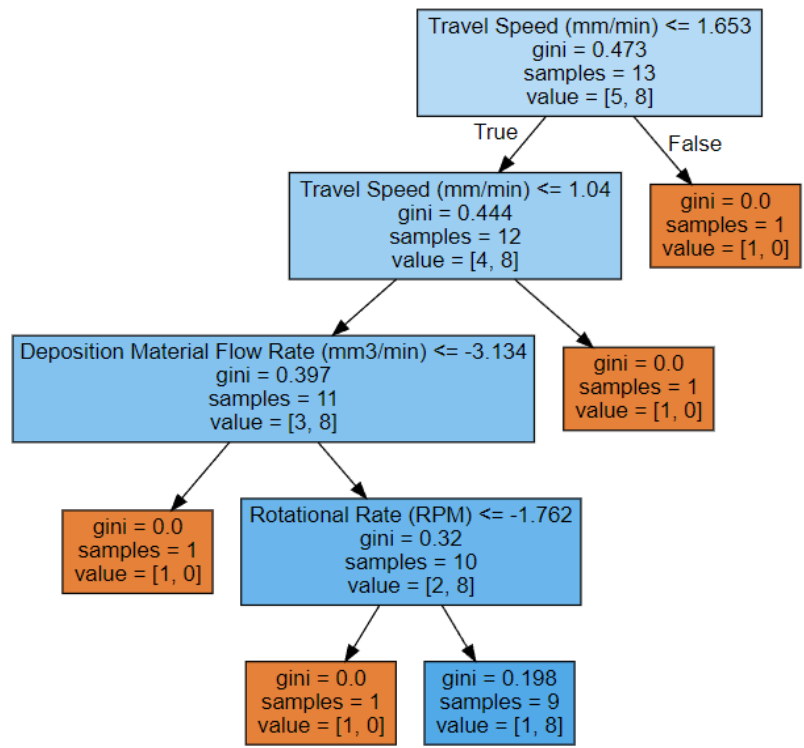

Figure 14. Decision tree plot obtained for classification based model

Similarly, Random Forest's mathematical formulation consists mostly of calculating impurity measurements, picking the optimal feature to partition the data, and merging the predictions from separate trees. To reduce overfitting, bootstrapped samples and random feature subsets are used. Random Forest Classification is a versatile and durable algorithm that is noted for its excellent accuracy and resistance to overfitting.

Gradient Boosting and Stochastic Gradient Boosting are both powerful strategies for creating extremely accurate models. They are frequently utilized in machine learning contests as well as real-world applications. The primary distinction is in how data is subsampled and the learning rate is modified. Because of its stochastic nature, stochastic gradient boosting is often preferred when dealing with big datasets or when computer resources are restricted.

Table 3 shows the obtained metric features for the classification based algorithms. Figure 15 shows the confusion matrix plot for the implemented classification based algorithms while Figure 16 shows the plot for Receiver Operating characteristics plots for each implemented algorithms.

Table 3. Results obtained from the supervised classification-based machine learning algorithms to predict deposition quality

| Algorithms | Execution Time in seconds | Training data accuracy | Test data accuracy | Overall F1-Score | ROC-AUC Score |
|---|---|---|---|---|---|
| Logistic | 0.0203 | 0.9230 | 1.0 | 1.0 | 1.0 |
| K-Nearest Neighbours | 0.0060 | 0.6153 | 0.5 | 0.5 | 1.0 |

| Support Vector | 0.5860 | 1.0 | 0.75 | 0.75 | 1.0 |
| Stochastic Gradient Descent | 4.4190 | 1.0 | 1.0 | 1.0 | 1.0 |
| Decision Tree | 0.0012 | 0.9230 | 0.5 | 0.5 | 0.5 |
| Random Forest | 0.3400 | 0.9230 | 0.75 | 0.75 | 0.75 |
| AdaBoost | 0.0100 | 1.0 | 1.0 | 1.0 | 1.0 |
| Gradient Boosting | 0.1300 | 1.0 | 1.0 | 1.0 | 1.0 |
| Stochastic Gradient Boosting | 0.3645 | 1.0 | 1.0 | 1.0 | 1.0 |

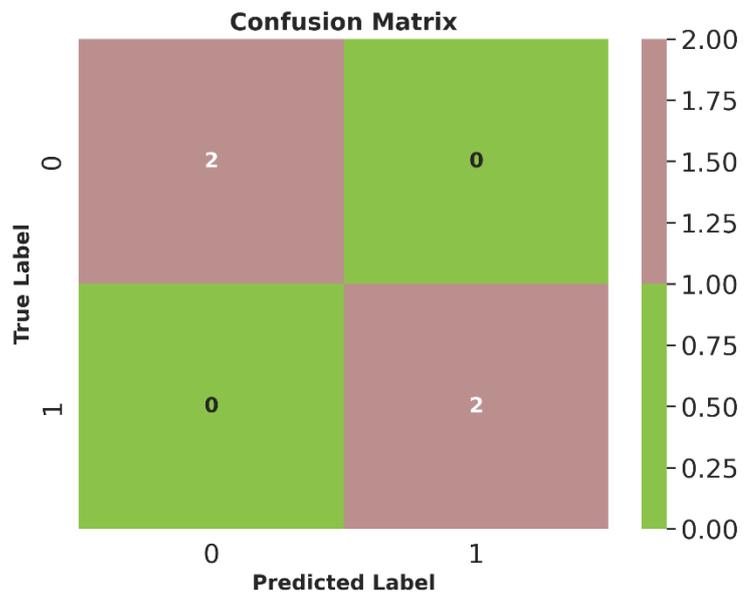

a)

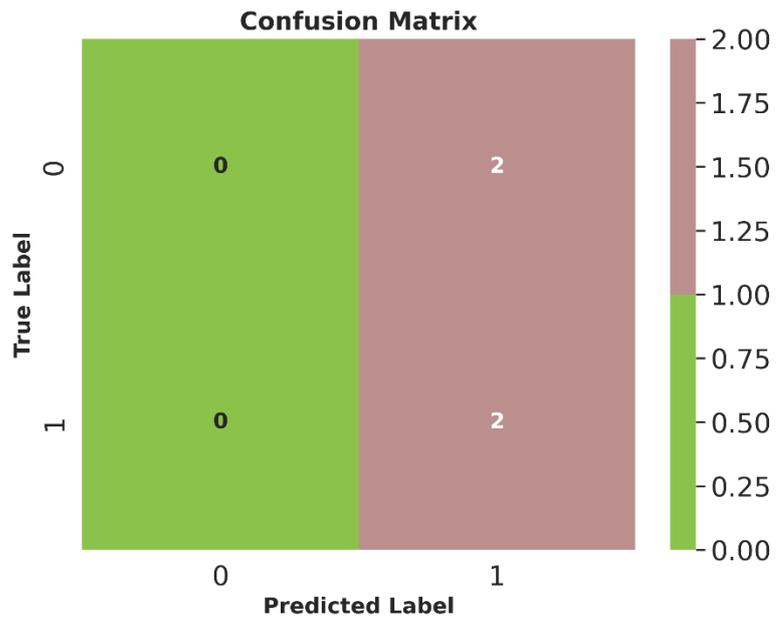

b)

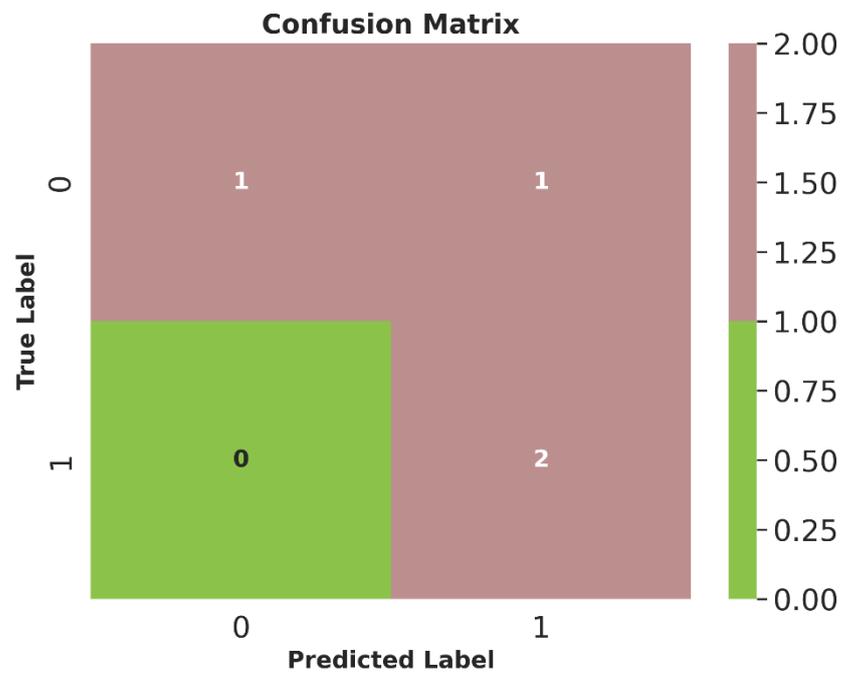

c)

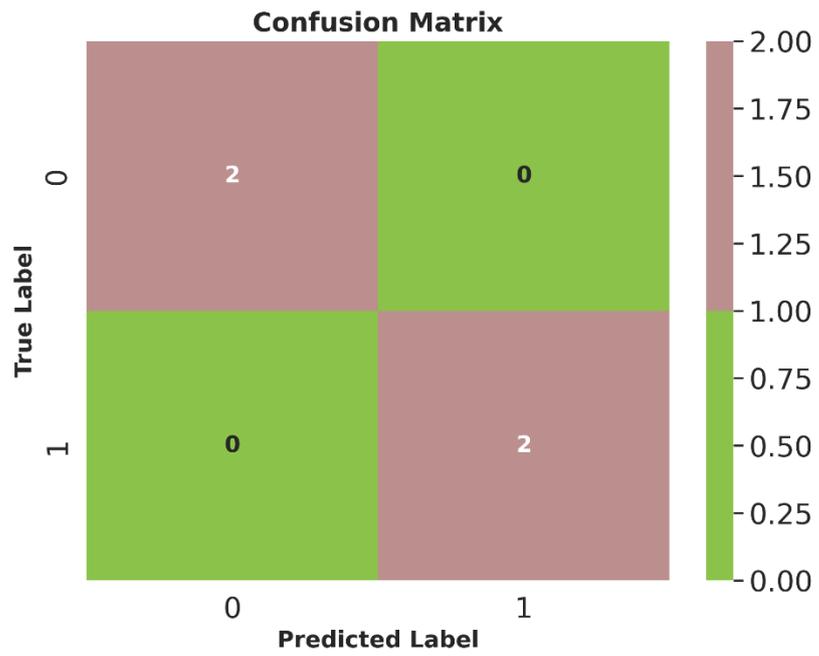

d)

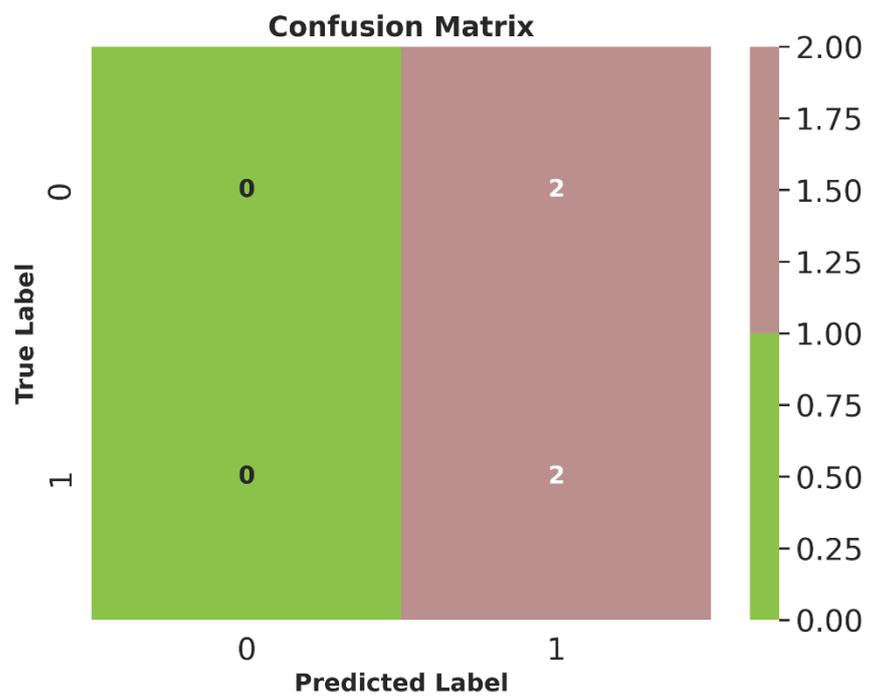

e)

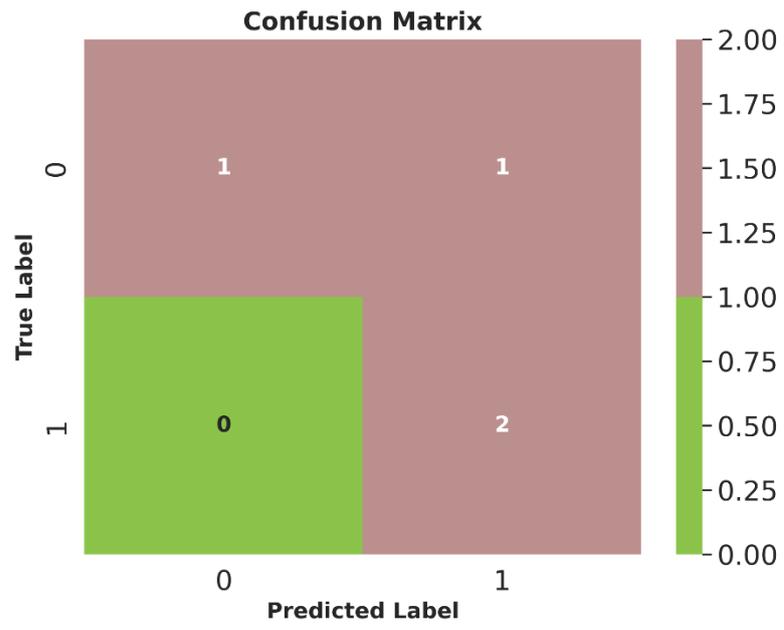

f)

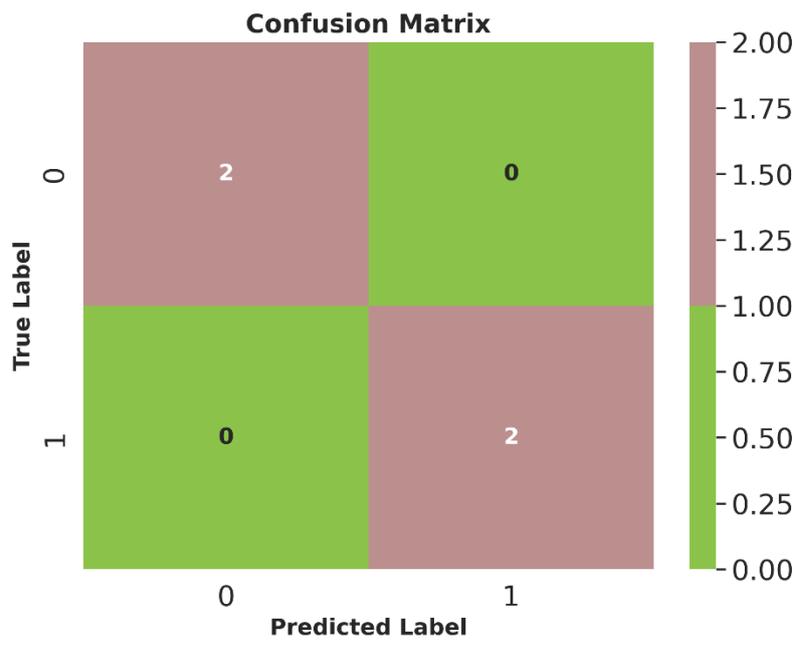

g)

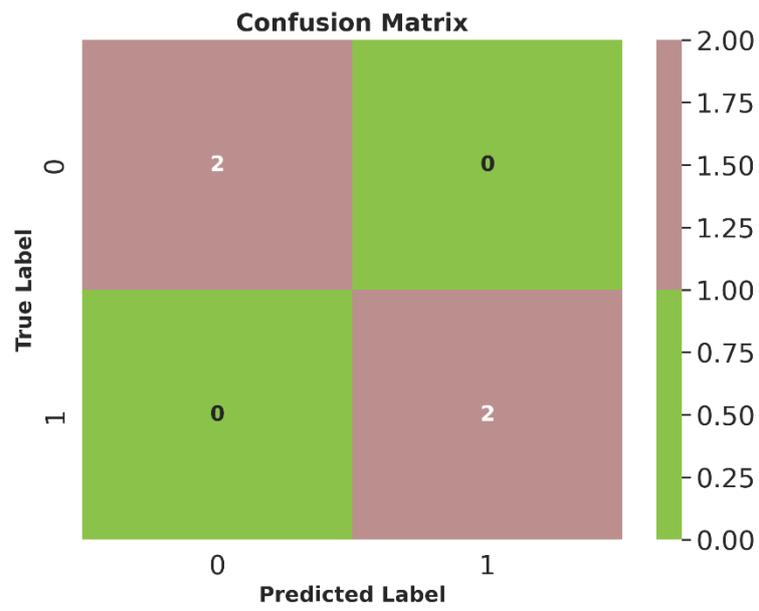

h)

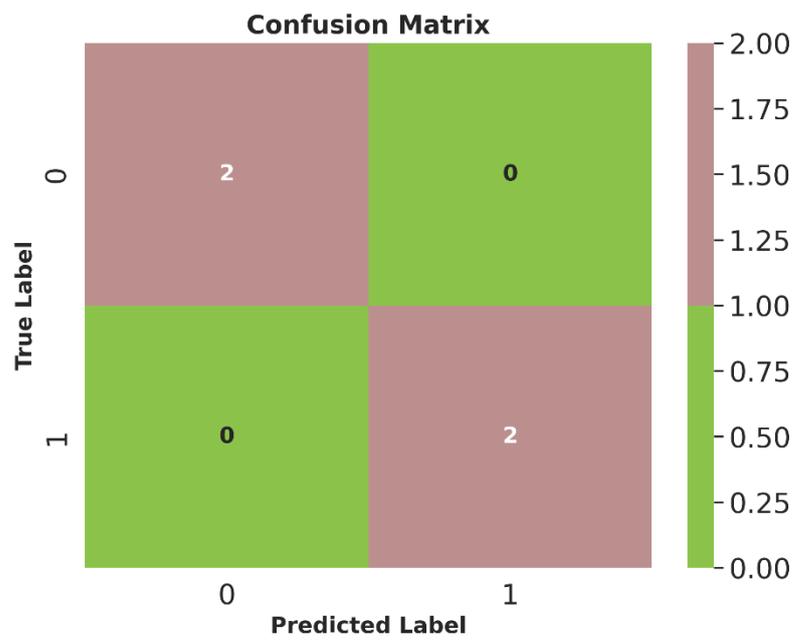

i)

Figure 15. Confusion matrix plots obtained for a) Logistic, b) K-Nearest Neighbours, c) Support Vector, d) Stochastic Gradient Descent, e) Decision Tree, f) Random Forest, g) AdaBoost, h) Gradient Boosting and i) Stochastic Gradient Boosting algorithms

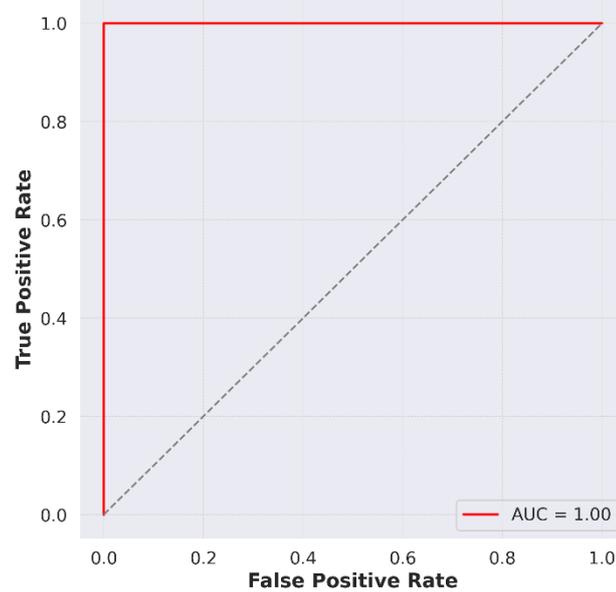

a)

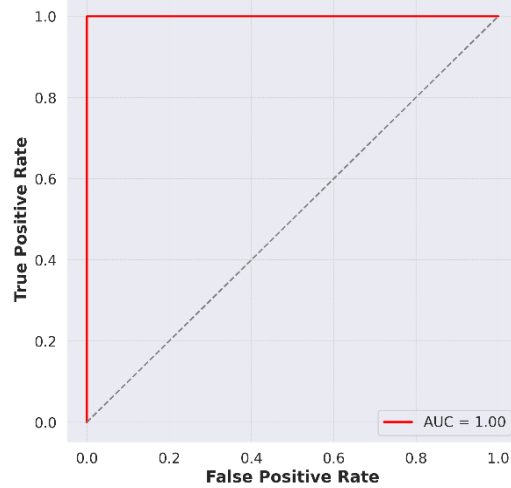

b)

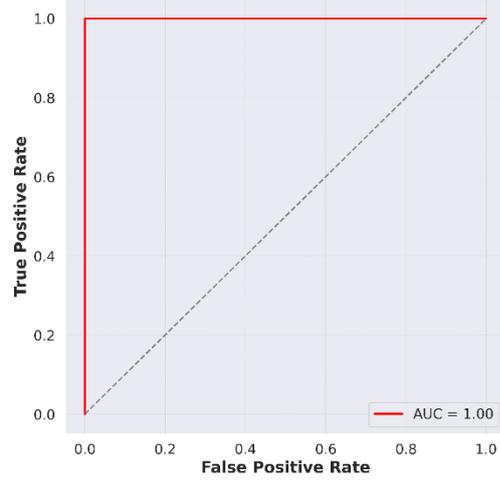

c)

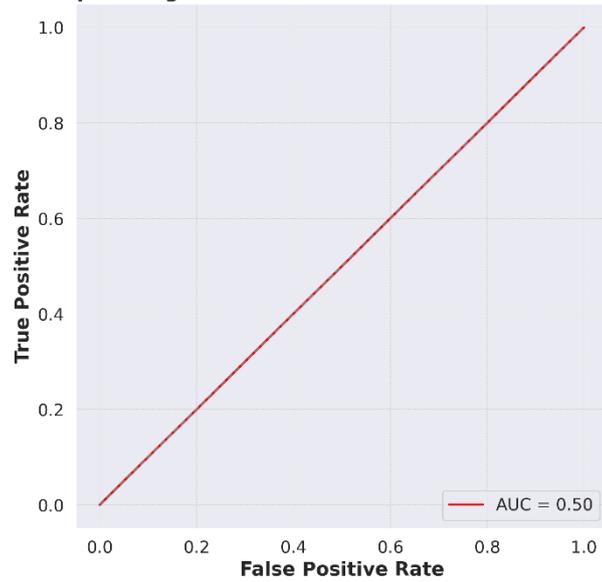

d)

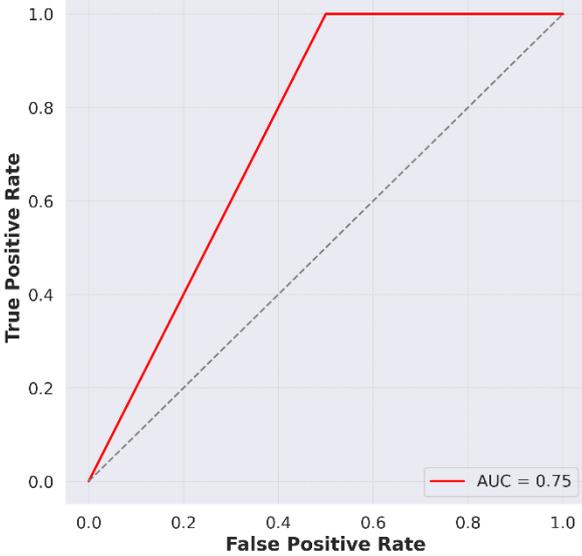

e)

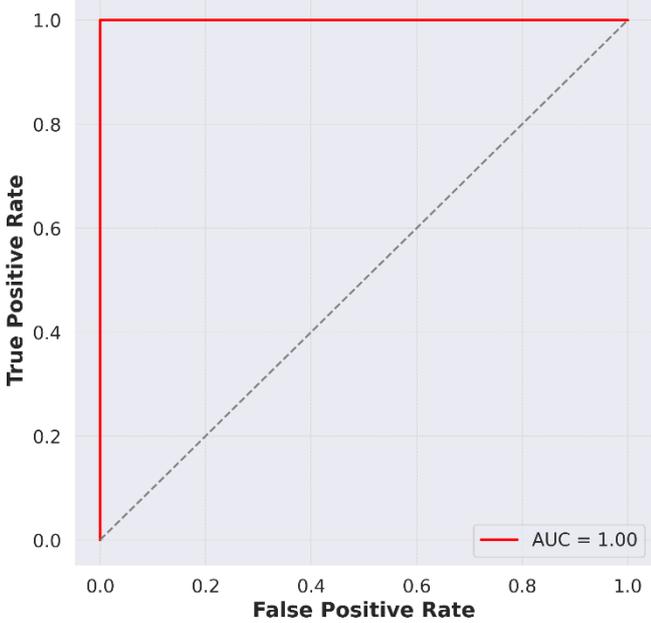

f)

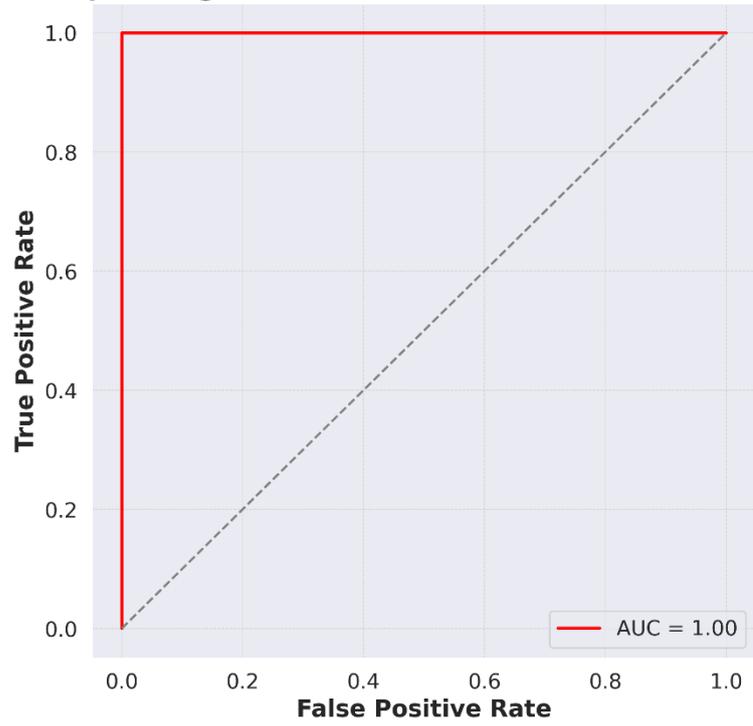

g)

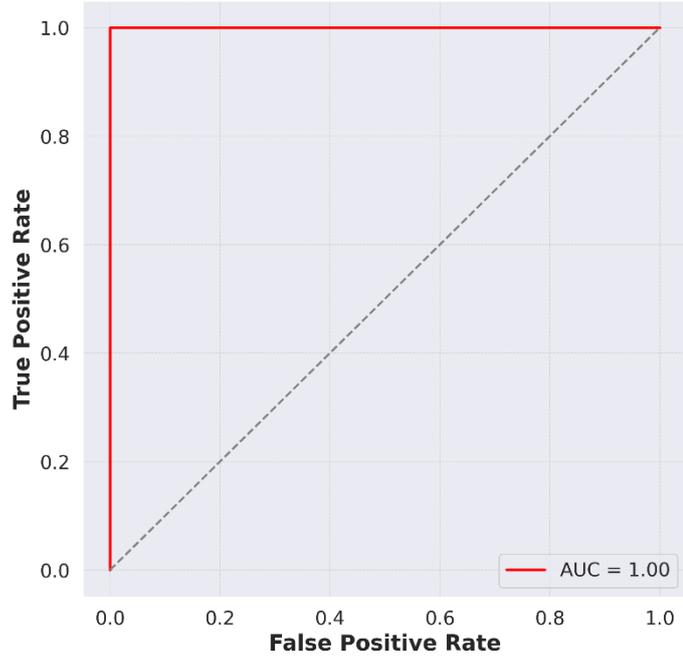

h)

Figure 16. ROC plots obtained for a) Logistic, b) K-Nearest Neighbours, c) Support Vector, d) Stochastic Gradient Descent, e) Decision Tree, f) Random Forest, g) AdaBoost, h) Gradient Boosting and i) Stochastic Gradient Boosting algorithms

Based on the results in Table 3, Figure 15, and Figure 16 for the classification algorithms predicting deposition quality it is observed that Logistic regression achieved perfect accuracy on both the training and test data with an ROC-AUC score of 1.0. The confusion matrix shows it correctly classified all good and poor depositions. This indicates logistic regression fit the data very well. K-nearest neighbors had low accuracy on the test data (50%) and training data (61.53%). The confusion matrix shows it struggled to differentiate between good and poor depositions. However, it still achieved an ROC-AUC score of 1.0, suggesting potential with tuning. Support vector classifier had 100% training accuracy but only 75% test accuracy. The confusion matrix shows it incorrectly classified some poor depositions as good. Its ROC-AUC score of 1.0 indicates good discrimination when tuning the decision threshold. Stochastic gradient descent matched logistic regression with 100% accuracy on both training and testing. The confusion matrix shows perfect classification. The ROC-AUC of 1.0 also indicates excellent performance. Decision tree had high training accuracy (92.3%) but low test accuracy (50%). The confusion matrix reveals overfitting, as it struggled on the unseen test data. The poor ROC-AUC score of 0.5 confirms this. Random forest improved upon decision tree with 75% test accuracy and ROC-AUC of 0.75. The confusion matrix shows it correctly classified more good and poor depositions than decision tree. AdaBoost, gradient boosting, and stochastic gradient boosting all achieved 100% training and testing accuracy with ROC-AUC of 1.0. Their confusion matrices show perfect classification of all samples.

## 5. Conclusion

This work demonstrated a novel integration of supervised machine learning and physics-informed neural networks to model peak temperature and deposition quality in additive friction stir deposition processes. Across several statistical measures, ensemble methods like gradient boosting and CatBoost proved most effective for regression-based peak temperature prediction within the supervised learning models. However, physics-informed models leveraging governing transport, wave, and heat equations significantly outperformed the data-driven approaches, achieving lowest errors by incorporating physical constraints. For classifying deposition quality, techniques like logistic regression and stochastic gradient descent delivered robust accuracy. The dual framework combining statistical and physics-based modeling provides unique insights into correlating process parameters to thermal profiles and deposition performance in AFSD. By elucidating these relationships, the integrated approach facilitates optimized design of AFSD processes for tailored microstructural properties. More broadly, this work highlights the merits of synergistically blending data-driven and physics-based techniques to uncover engineering design principles linking manufacturing processes to materials structure and properties.

**Funding Information:** No external funding was received for this research work.

**Competing Interest Statement:** Author declare no competing interests in the present work.